\documentclass[11pt]{article}

\usepackage[margin=1in]{geometry}
\usepackage[T1]{fontenc}
\usepackage{amsmath}
\usepackage{newtxtext,newtxmath}
\usepackage{graphicx}
\usepackage{booktabs}
\usepackage{microtype}
\usepackage[round]{natbib}
\usepackage{authblk}
\usepackage[hidelinks]{hyperref}

\newcommand{\Description}[1]{}

\hypersetup{
  pdftitle={There Is No Neutral Harness: Modern LLM Leaderboards Are Manufactured by Config-Fragile Items},
  pdfauthor={V.S. Raghu Parupudi}
}

\title{There Is No Neutral Harness:\\[2pt] Modern LLM Leaderboards Are Manufactured by Config-Fragile Items}
\author{V.S. Raghu Parupudi}
\affil{University of California, San Diego}
\affil{\texttt{pvsrrkishore@gmail.com}}
\date{\today}

\begin{document}

\maketitle

\begin{abstract}
A multiple-choice benchmark fixes the questions and the correct answers. It does not fix the order of the options, the wording of the prompt, or whether a language model's answer is read from generated text or from per-option likelihoods. Work on this sensitivity measures it on the aggregate score or the aggregate rank and reports it as variance to be controlled. That leaves the item level unexamined, so it is not known which items the variance falls on, or whether they are the items that separate one model from the next. We treat the evaluation harness as an independent variable and resolve its effect down to single items. We introduce the \textit{fragility grid}, in which 12 open-weight instruction-tuned language models from 4 families answer the same 3{,}679 items from 4 benchmarks under 26 equally defensible harness configurations, recording one correctness bit for every model, item, and configuration. The comparison is matched, since the items, the weights, and the greedy decoding stay fixed while only the harness varies, and no configuration in the grid is adversarial. Under the grid a model's score is a band rather than a point. gemma4-31b, with fixed weights and fixed items, scores between 31 and 89 percent depending only on the harness. Three results follow. On the items that two adjacent models both answer stably, the pair is tied. Five of the 11 pairs answer identically and four more differ on a single item, yet every gap that exists on the full set is ordered anyway, and config-fragile items carry 95.7 percent of a pair's gap on average. Four of the 12 models reach rank one under some configuration and 8 reach the top three, so the choice of harness selects the winner. Item discrimination, the property that benchmark-compression methods maximize, correlates with fragility at 0.28 (95 percent CI 0.25 to 0.30), and the most discriminative items are config-fragile for 96 percent of models against 85 percent across the benchmark, so compression keeps the fragile items rather than removing them. The scoring choice, not the option order that protocols usually fix and not the prompt wording, is the load-bearing axis. We release the per-item records and the analysis script, from which every number in the paper regenerates on a CPU in seconds, and we position the fragility grid as a check that a leaderboard can run before it reports an order.
\end{abstract}


\section{Introduction}

A benchmark score is reported as a property of a model. Practitioners read MMLU or ARC accuracy as a fact about the system, compare two numbers, and treat the larger one as the better model. The number is produced by a procedure with free parameters that the benchmark does not fix. The options can be labeled with letters or digits. They can be shown in their original order or shuffled. The answer can be parsed from generated text or selected as the highest-likelihood option. Each choice is defensible, and each appears in evaluation code that people use. That such choices move the score is established, and whether these benchmarks measure what their scores are taken to mean is itself contested \citep{bowman2021benchmarking}. Existing work on the sensitivity differs in what it perturbs, covering prompt formatting \citep{sclar2024formatting}, instruction paraphrase \citep{mizrahi2024multiprompt}, option order \citep{pezeshkpour2024order}, and whole evaluation setups \citep{alzahrani2024targets}. These lines share a structural property. Each measures the sensitivity on an aggregate, the score or the rank, and reports it as variance to be pinned down with a fixed protocol.

Aggregate variance is enough when the question is whether a score is noisy. It is not enough when the question is whether an ordering is real. A noisy score whose noise falls evenly across the benchmark is a different object from a noisy score whose noise falls precisely on the items that separate two neighboring models. The two are indistinguishable at the aggregate level and have opposite consequences for a leaderboard. Resolving them requires descending to the item. \textit{We ask: when the harness changes the ranking, which items carry the gap between two models, and does the ordering survive when those items are set aside?}

We answer this by promoting the harness from an implicit choice to a controlled variable. The instrument is a \textit{fragility grid}:\footnote{Code, per-item records, and analysis script: \url{https://github.com/NikolaTesla-007/fragility-grid}.} a set of 26 harness configurations built from three axes that any evaluator picks between, namely option order, prompt format, and scoring method. Every model answers every item under every configuration, which yields a correctness bit for each triple of model, item, and configuration. The grid is deliberately plain. It introduces no new method and no new benchmark, and its only job is to hold the items, the weights, and the decoding fixed so that the harness is the one thing that moves. What the instrument buys is a partition of the benchmark. An item is \textit{robust} for a model when its correctness is the same under all 26 configurations, and \textit{config-fragile} when it flips. With that partition, each gap on the leaderboard splits into the part supported by items both models answer stably and the part carried by items that flip.

The split is lopsided. Figure~\ref{fig:band} shows the first consequence, that a model's accuracy is a band rather than a point. gemma4-31b, with the same weights and the same items, scores 31 percent under one configuration and 89 percent under another. Across the roster, 85 percent of the answers a model gets credit for can be flipped to wrong by an equally valid configuration. The second consequence is the one that matters for ranking. On the items that two neighboring models both answer stably, the models are tied. The ordering that the leaderboard prints is produced by the items they are least sure about.

Our contributions are the following.

\begin{itemize}
\item \textbf{Adjacent-model orderings are not identified by the stable items.} On the items that both models in an adjacent pair answer stably, 5 of the 11 pairs answer identically item for item, 4 differ on exactly one item out of the 75 to 158 available, and the remaining 2 disagree in both directions and net to zero. Every one of the 10 pairs that has a nonzero gap on the full set is ordered anyway, and config-fragile items carry 95.7 percent of a pair's gap on average, rising to 99.7 percent when the items are pooled rather than the pairs averaged. To our knowledge, this is the first decomposition of a leaderboard gap into robust and fragile mass.
\item \textbf{Choosing a harness is choosing a winner.} Four of 12 models reach rank one under some configuration in the grid and 8 of 12 reach the top three. mixtral-8x7b places 11th of 12 under the reference harness and first under another. This is an existence statement about the set of defensible champions rather than a measurement of how far ranks drift. We attach the margin behind every first place: the two models that take 24 of the 26 configurations win by median leads of 181.5 and 98.5 items out of 3{,}679, and the other two top a single configuration each, by 1 item and by 3 items.
\item \textbf{Benchmark compression keeps the fragile items.} Item discrimination, the quantity that item-response-theory compression maximizes, correlates with config-fragility at 0.28 (95 percent CI 0.25 to 0.30), and the most discriminative items have mean fragility 0.96 against 0.85 across the benchmark. Compressing to 100 items raises the champion count from 4 on the full benchmark to 5.1 for a random subset of that size, and selecting the 100 items by discrimination raises it further to a mean of 5.7, with a range of 4 to 7 across seeded tie-breaks of an ill-defined cutoff. The correlation and the fragility gap do not depend on that tie-break, and we rest the claim on them.
\item \textbf{The scoring axis is load-bearing.} Holding the other axes at the reference, mean accuracy robust to the scoring choice is 0.31, below prompt format at 0.40 and option order at 0.60. The generation-versus-likelihood choice, not the option shuffling that protocols usually fix, destroys the most credited answers.
\end{itemize}

\section{Related Work}

\textbf{Sensitivity of scores to presentation.} A well-developed line shows that language model scores depend on how a task is presented. \citet{sclar2024formatting} vary prompt formatting in ways that preserve meaning and find accuracy swings wide enough to reorder systems, with no format uniformly best. \citet{mizrahi2024multiprompt} run many paraphrased instructions per task and show that one prompt gives a single point from a distribution that should be reported in full. For multiple choice, \citet{pezeshkpour2024order} document sensitivity to option order and connect it to items where a model is uncertain between its leading choices, which is the mechanism our results build on. A related line traces the cause to position and label biases in the prediction and to the order of in-context examples \citep{zhao2021calibrate, lu2022fantastically}. These studies establish that the score moves, and they measure the movement on the aggregate. Which items move, and whether those items are the ones that separate two models, is left open.

\textbf{Leaderboard perturbation and rank movement.} Closest in spirit is \citet{alzahrani2024targets}, who perturb the evaluation setup behind a public leaderboard and report that model ranks travel by several positions. Holistic evaluation makes the complementary argument that one setup gives a partial view and that many conditions should be reported together \citep{liang2023helm}, and preference-based leaderboards face their own aggregation questions \citep{chiang2024arena}. These works report how far ranks travel, which is a dispersion statistic. A dispersion statistic does not say whether a particular ordering is supported by the data, and it does not say how many distinct models could each be a defensible winner. Whether a reported gap is large enough to be real is a question of statistical power, which is often lacking in language-model comparisons \citep{card2020power}, and our robust-item test is one answer to it.

\textbf{Scoring method as an undocumented choice.} Whether an answer is generated and parsed or selected by likelihood is itself a harness decision, and it changes which model looks stronger \citep{robinson2023mcqa}. The likelihood of an option competes with the raw frequency of its surface form, so the highest-probability option is not always the intended answer \citep{holtzman2021surface}, yet per-option probabilities stay informative about correctness even when they are miscalibrated \citep{plaut2024softmax}. Neither path can be dismissed as reading noise. The likelihood path is the default in widely used evaluation code \citep{eval-harness}, while generation with answer parsing is standard for chat-style systems, and papers frequently omit which was used. This line establishes the scoring choice as consequential. None of it ranks the scoring choice against the other harness axes on a common item set, which is what our axis decomposition does.

\textbf{Benchmark compression.} A productive line compresses benchmarks to a small set of informative items using item response theory, showing that on the order of a hundred well-chosen items reproduce full-benchmark scores and ranks \citep{polo2024tinybenchmarks, embretson2000irt}. Item response theory has been applied to compare test sets and to reweight leaderboard examples by how much they discriminate between systems \citep{vania2021comparing, rodriguez2021evaluation}, and efficient-benchmarking work selects informative subsets to cut evaluation cost \citep{perlitz2024efficient}. The selection criterion is item discrimination, and the validation is performed under a fixed harness. Whether the retained items survive a change of harness is not tested. It is precisely the retained items that we find to be the fragile ones, which makes the compressed benchmark less stable across configurations than the benchmark it replaces.

\textbf{Our position, and what we do not claim.} Across these lines the harness enters as a source of variance on an aggregate score, to be reported or minimized, and the variance is never resolved to the items it acts on. That resolution is what the fragility grid adds, and the results follow from it: the gap decomposition, the champion count, the link from compression to fragility, and the ranking of the harness axes against one another. The dangerous reading of this work should be disowned plainly. We do not claim that these models are equally capable, and we do not claim that multiple-choice benchmarks carry no signal about capability. We claim only that the \textit{order} of adjacent models under a single harness is not identified by the items those models answer stably, and that the identity of the top model is a function of a configuration choice that is rarely reported.

\section{The Fragility Grid}

\subsection{Models}
The roster is 12 instruction-tuned models from 4 families, spanning 3 billion to 70 billion parameters and including both dense and mixture-of-expert architectures. Qwen contributes qwen3-4b, qwen3-14b, qwen3-32b, and the mixture-of-expert qwen3-30b-a3b \citep{qwen3}. Llama contributes llama3.2-3b, llama3.1-8b, and llama3.3-70b \citep{dubey2024llama3}. Gemma contributes gemma4-e4b, gemma4-12b, gemma4-31b, and the mixture-of-expert gemma4-26b-a4b \citep{gemma3}. Mistral is represented by mixtral-8x7b \citep{jiang2024mixtral}. The spread of families and scales is deliberate, so that a result holding across all 12 cannot be a quirk of one training recipe or one size. Two further Mistral models were prepared and dropped after they hung during inference-engine initialization, which is an infrastructure failure and not a property of their answers.

\subsection{Benchmarks and items}
We use four standard multiple-choice benchmarks: ARC \citep{clark2018arc}, HellaSwag \citep{zellers2019hellaswag}, MMLU \citep{hendrycks2021mmlu}, and TruthfulQA \citep{lin2022truthfulqa}. Up to 1{,}000 items are drawn from each with a fixed seed. TruthfulQA supplies 679 after four-option filtering, so the pooled set is 3{,}679 items per model, and every item has four options and one correct answer. All 12 models see the identical item set, so any difference between models is a difference in answers and not in questions. TruthfulQA is not natively four-option, and the loader builds each item from the correct answer plus the first three distractors, which places the gold answer first in the benchmark's own order. The option-order axis moves it in 5 of the 6 orderings, but the reference configuration inherits it, so on those 679 items the reference answer key is uniformly the first option. Appendix~\ref{app:items} gives the construction and Appendix~\ref{app:loo} reports every headline number recomputed with TruthfulQA removed.

\subsection{The grid}
A \textit{configuration} is a triple of choices that a real evaluator makes and can defend. Three axes vary across the grid.

\begin{description}
\item[Option order.] Six orderings of the four options, namely the original order and five fixed shuffles. Only the positions change, and the text of the options is untouched.
\item[Prompt format.] Four generation formats that differ only in surface presentation: letters with periods, letters in parentheses, digit labels, and an instruction that asks explicitly for the answer letter. Two cloze stems scored by likelihood are order invariant and complete the format axis.
\item[Scoring method.] Generation, where the emitted answer is parsed, against likelihood, where the option with the highest length-normalized continuation log-probability wins. The likelihood path is the default in common evaluation code.
\end{description}

Four generation formats crossed with six orderings give 24 configurations, and the two likelihood cloze stems bring the total to 26. One configuration is designated the \textit{reference}: letter labels, original order, generation scoring. It defines the leaderboard that everything else is measured against. Every configuration in the grid appears in published evaluation setups, and none was constructed to break a model, so the grid is not adversarial. That is the point of it. A grid of adversarial harnesses would show only that a determined attacker can move a score, whereas a grid of defensible harnesses shows that an honest evaluator can.

\subsection{Definitions}
For each model and item we hold a correctness bit under all 26 configurations. Everything below is computed from that tensor.

\begin{description}
\item[Robust and fragile items.] An item is \textit{robust-correct} for a model when it is correct under all 26 configurations, \textit{robust-wrong} when it is wrong under all 26, and \textit{config-fragile} when its correctness flips. Robust-wrong items matter: they keep uniformly-failed items from being confused with fragile ones.
\item[Accuracy band.] Robust accuracy counts items correct under every configuration, reference accuracy counts items correct under the reference, and optimistic accuracy counts items correct under at least one. Robust accuracy is at most reference accuracy, which is at most optimistic accuracy.
\item[Config-lucky fraction.] The share of a model's reference-correct answers that flip to wrong under some other valid configuration.
\item[Fragile share of gap.] For an adjacent pair on the reference leaderboard, the fraction of the reference accuracy gap attributable to config-fragile items rather than to items both models answer robustly.
\item[Order manufactured.] An adjacent pair is \textit{manufactured} when the two models are statistically tied on their jointly robust items, meaning a bootstrap confidence interval on the robust-item accuracy difference contains zero, yet are separated on the full item set.
\item[Champion set.] The set of models that attain rank one under at least one configuration in the grid.
\end{description}

\subsection{Controls and inference}
The comparison is matched by construction. The items, the model weights, and the decoding are held fixed across the grid, and only the harness varies, so a change in correctness cannot be attributed to a different question, a different checkpoint, or a different sampling draw. Decoding is greedy at temperature zero and zero-shot throughout, which matches how static leaderboards are scored and removes sampling as a competing explanation. Confidence intervals on accuracy differences are bootstrap intervals over items \citep{efron1994bootstrap}, a nonparametric choice suited to the paired, per-item structure of the comparison \citep{dror2018hitchhiker}. Inference runs under vLLM \citep{kwon2023vllm} on four 48-gigabyte GPUs. Every randomized step, which means the bootstrap resamples, the random item subsets, and the tie-breaks at the discrimination cutoff, is drawn from a single seeded generator, so the analysis is a deterministic function of one seed and two runs give byte-identical output. We release the per-item records, which are 48 files holding 26 correctness bits per item per model, along with the analysis script, at \url{https://github.com/NikolaTesla-007/fragility-grid}, so that every number in this paper regenerates on a CPU in about ten seconds. Appendix~\ref{app:repro} states the item selection, the 26 templates and orderings, the decoding parameters, and every step that consumes randomness, at the level of detail needed to rebuild the grid from scratch rather than only to rerun our code.

\section{Results}

\subsection{A model's accuracy is a band}
\begin{figure}[ht]
  \centering
  \includegraphics[width=0.85\linewidth]{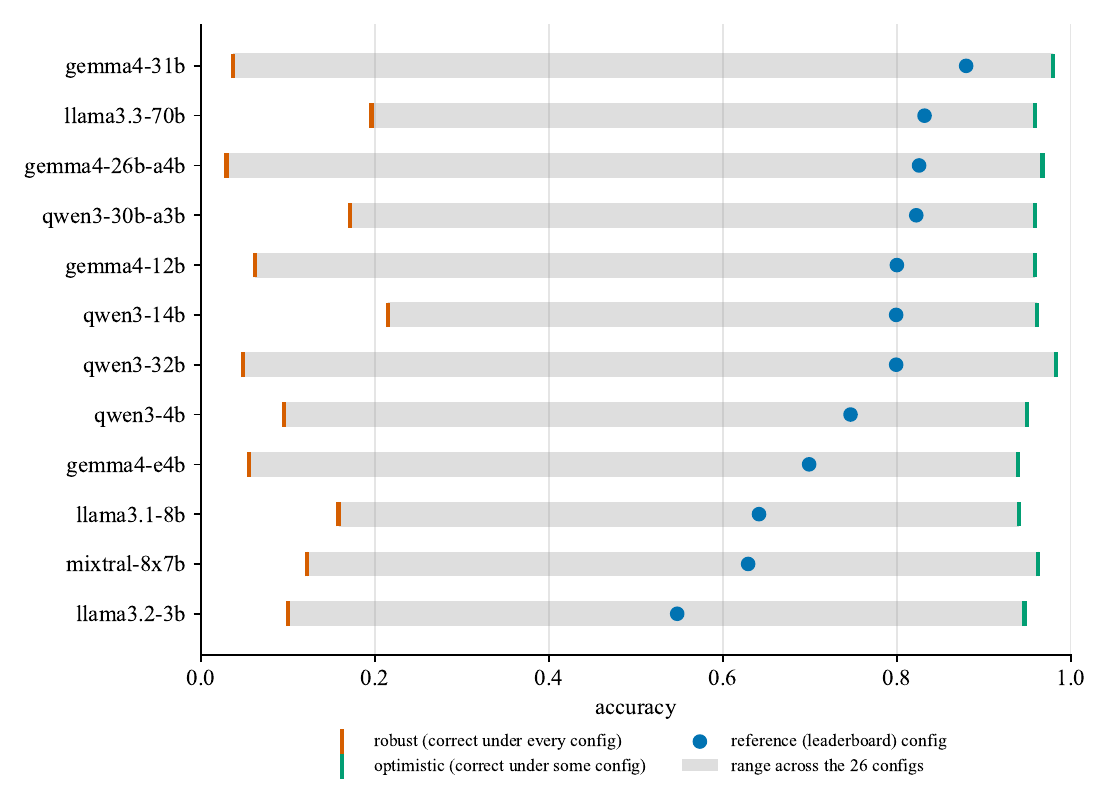}
  \caption{The same model, the same items, and many scores. Models run down the axis in reference-leaderboard order, best at the top. Each row runs from robust accuracy, the items a model answers correctly under all 26 harness configurations (left tick), to optimistic accuracy, the items it answers correctly under at least one (right tick), with the reference-configuration score marked by the dot. Only the harness differs along a row. For gemma4-31b the row runs from 0.04 to 0.98, and the 0.88 it is credited with under the reference harness is one point inside that range. Averaged across the 12 models, 85 percent of credited answers can be flipped to wrong by an equally defensible configuration, which is why a single leaderboard number is a point drawn from a wide band rather than a property of the model.}
  \Description{A horizontal chart with one row per model, ordered by reference accuracy with the best model at the top. Accuracy runs along the horizontal axis. For each model a shaded bar runs from a low robust-accuracy value on the left to a high optimistic-accuracy value on the right, with a dot marking the reference accuracy. The reference dots sit far to the right of the left ends, showing that only a small fraction of items are answered correctly under every configuration while a large fraction are answered correctly under at least one.}
  \label{fig:band}
\end{figure}

Table~\ref{tab:leaderboard} gives the reference leaderboard together with the lowest and highest single-configuration accuracy each model reaches, and Figure~\ref{fig:band} draws the wider robust-to-optimistic band for the same roster.

\begin{table}
  \caption{The reference leaderboard, with the lowest and highest single-configuration accuracy across the 26 configurations. Spread is their difference, computed before rounding, so for two rows it differs by 0.001 from the difference of the rounded columns shown. Config-lucky is the share of reference-correct answers that flip to wrong under some valid configuration. Accuracy is pooled over the same 3{,}679 items for every model. The spread reported here is the range of single-configuration scores and is narrower than the robust-to-optimistic band drawn in Figure~\ref{fig:band}.}
  \label{tab:leaderboard}
  \begin{tabular}{@{}rlccccc@{}}
    \toprule
    Rank & Model & Reference & Min-config & Max-config & Spread & Config-lucky \\
    \midrule
    1  & gemma4-31b      & 0.880 & 0.306 & 0.886 & 0.580 & 0.958 \\
    2  & llama3.3-70b    & 0.832 & 0.520 & 0.832 & 0.311 & 0.764 \\
    3  & gemma4-26b-a4b  & 0.825 & 0.302 & 0.840 & 0.538 & 0.964 \\
    4  & qwen3-30b-a3b   & 0.822 & 0.514 & 0.835 & 0.321 & 0.791 \\
    5  & gemma4-12b      & 0.800 & 0.284 & 0.810 & 0.526 & 0.923 \\
    6  & qwen3-14b       & 0.799 & 0.559 & 0.804 & 0.245 & 0.731 \\
    7  & qwen3-32b       & 0.799 & 0.380 & 0.802 & 0.422 & 0.939 \\
    8  & qwen3-4b        & 0.747 & 0.398 & 0.754 & 0.356 & 0.873 \\
    9  & gemma4-e4b      & 0.699 & 0.289 & 0.705 & 0.416 & 0.921 \\
    10 & llama3.1-8b     & 0.641 & 0.458 & 0.645 & 0.186 & 0.753 \\
    11 & mixtral-8x7b    & 0.629 & 0.476 & 0.661 & 0.185 & 0.806 \\
    12 & llama3.2-3b     & 0.547 & 0.395 & 0.608 & 0.213 & 0.818 \\
    \bottomrule
  \end{tabular}
\end{table}

The mean config-lucky fraction across the 12 models is 0.85, so most of what a model is credited with is credited conditionally. Robust accuracy, the part that survives every configuration, is at most 0.215 for any model in the roster and below 0.07 for five of them. Sensitivity does not track leaderboard position. The widest configuration spread of all belongs to gemma4-31b at 0.580, which is the top-ranked model, and the narrowest belong to mixtral-8x7b and llama3.1-8b at 0.185 and 0.186. Instability is therefore not a symptom of being a weak model.

\subsection{The ordering rests on the fragile items}
\begin{figure}[ht]
  \centering
  \includegraphics[width=0.85\linewidth]{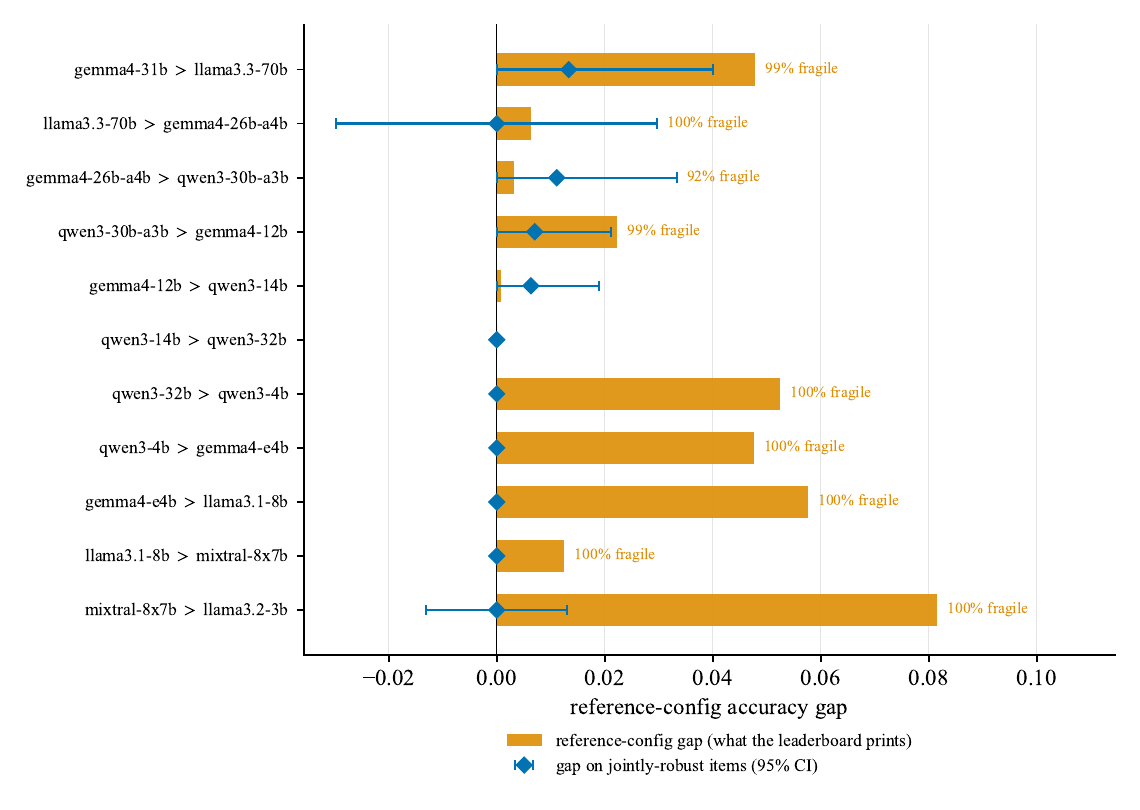}
  \caption{Every leaderboard gap is fragile mass. For each adjacent pair, the orange bar is the reference-config accuracy gap, the ordering the leaderboard prints. The blue marker is the accuracy gap on the items both models answer stably, each model's answer unchanged across all 26 configurations, with its 95 percent bootstrap confidence interval. Every marker sits on zero and every interval contains it, so the two models are tied once the fragile items are set aside, while the orange bar they are ranked by remains. The label on each bar gives the fragile share of that gap. The qwen3-14b and qwen3-32b pair has no gap to decompose, since the two are tied on the full item set as well.}
  \Description{A horizontal chart with one row per adjacent pair of models, in leaderboard order. Each row has a long orange bar giving the reference-configuration accuracy gap, and a blue diamond with a horizontal error bar giving the gap on jointly robust items. Every blue diamond sits at zero with an interval that crosses zero, while the orange bars extend well to the right, showing that the ranked gap lives entirely in the fragile items. The qwen3-14b versus qwen3-32b row has no orange bar because that pair has a zero gap.}
  \label{fig:gap}
\end{figure}

Table~\ref{tab:pairs} decomposes each adjacent-pair gap, and Figure~\ref{fig:gap} draws it.

\begin{table}
  \caption{Adjacent-pair gap decomposition. Reference gap is the accuracy difference under the reference configuration. Fragile share is the part of that gap carried by config-fragile items. Robust tie indicates that the two models are indistinguishable on the items both answer stably, where five pairs answer identically on every such item, four differ on exactly one item, and two disagree in both directions and net to zero. Robust items is the size of that jointly stable set. Reversing counts the configurations under which the lower model outscores the higher one.}
  \label{tab:pairs}
  \small
  \begin{tabular}{@{}llccccc@{}}
    \toprule
    Higher & Lower & Ref gap & Fragile share & Robust tie & Robust items & Reversing / 26 \\
    \midrule
    gemma4-31b     & llama3.3-70b   & 0.048 & 0.994 & yes & 75  & 8 \\
    llama3.3-70b   & gemma4-26b-a4b & 0.006 & 1.000 & yes & 101 & 10 \\
    gemma4-26b-a4b & qwen3-30b-a3b  & 0.003 & 0.917 & yes & 90  & 17 \\
    qwen3-30b-a3b  & gemma4-12b     & 0.022 & 0.988 & yes & 142 & 10 \\
    gemma4-12b     & qwen3-14b      & 0.001 & 0.667 & yes & 158 & 21 \\
    qwen3-14b      & qwen3-32b      & 0.000 & n/a   & yes & 151 & 10 \\
    qwen3-32b      & qwen3-4b       & 0.052 & 1.000 & yes & 72  & 1 \\
    qwen3-4b       & gemma4-e4b     & 0.048 & 1.000 & yes & 143 & 12 \\
    gemma4-e4b     & llama3.1-8b    & 0.058 & 1.000 & yes & 193 & 8 \\
    llama3.1-8b    & mixtral-8x7b   & 0.013 & 1.000 & yes & 336 & 19 \\
    mixtral-8x7b   & llama3.2-3b    & 0.082 & 1.000 & yes & 230 & 0 \\
    \bottomrule
  \end{tabular}
\end{table}

The result is uniform across the leaderboard, and it is best stated by counting items rather than by quoting an interval. On the items that two adjacent models both answer stably, five of the eleven pairs answer \textit{identically}, item for item, with no disagreement anywhere. Four more pairs disagree on exactly one item out of the 75 to 158 they both answer stably. The remaining two disagree in both directions and net to zero. A bootstrap confidence interval on the robust-item accuracy difference contains zero for all 11 pairs, but we report the item counts as the primary evidence because the interval is the weaker instrument here. For the nine pairs that have no disagreements in one direction, the lower bound of the interval sits at zero by construction rather than by evidence, so containing zero is not something those pairs could have failed to do. The item counts carry the claim without that difficulty.

Ten of the pairs have a nonzero gap on the full item set, and all 10 of those are manufactured, meaning the ordering exists on the full set and vanishes on the robust set. The 11th pair, qwen3-14b against qwen3-32b, is a genuine null: the two models are tied on the full set as well, with a reference gap of exactly 0.000, and we report it as a finding rather than an exception. Averaged over the 10 pairs that have a gap, 95.7 percent of that gap is fragile mass. That average is unweighted, so the one pair whose entire gap is three items contributes as much as the pair whose gap is 300. Pooling the items instead of averaging the pairs raises the fragile share to 99.7 percent, and we report the smaller number.

The instability is not confined to distant neighbors. The gemma4-12b and qwen3-14b pair reverses under 21 of 26 configurations, and llama3.1-8b and mixtral-8x7b reverse under 19. Taking the whole ranking at once, the Kendall correlation between the reference ordering and each other configuration's ordering averages 0.42, so the typical configuration flips about 29 percent of the 66 pairwise model orderings. Its minimum is $-0.09$, at which the ordering carries no information about the reference one, with marginally more pairs reversed than preserved.

\textbf{A natural worry.} Fragile items might simply be hard items, in which case a leaderboard decided by hard items would be working as intended. Two things in the data speak against that reading. Robust-wrong items, the ones every configuration gets wrong, form their own class and sit inside the jointly robust set, so uniformly hard items are counted on the robust side of the split rather than the fragile side. More directly, fragility tracks item \textit{discrimination} rather than difficulty, as the next section shows, and discrimination is exactly the property that makes an item separate strong models from weak ones. The items that carry the gap are the items built to carry the gap, and those are the items the harness moves.

\subsection{Choosing a harness is choosing a winner}
\label{sec:champions}
\begin{figure}[ht]
  \centering
  \includegraphics[width=0.85\linewidth]{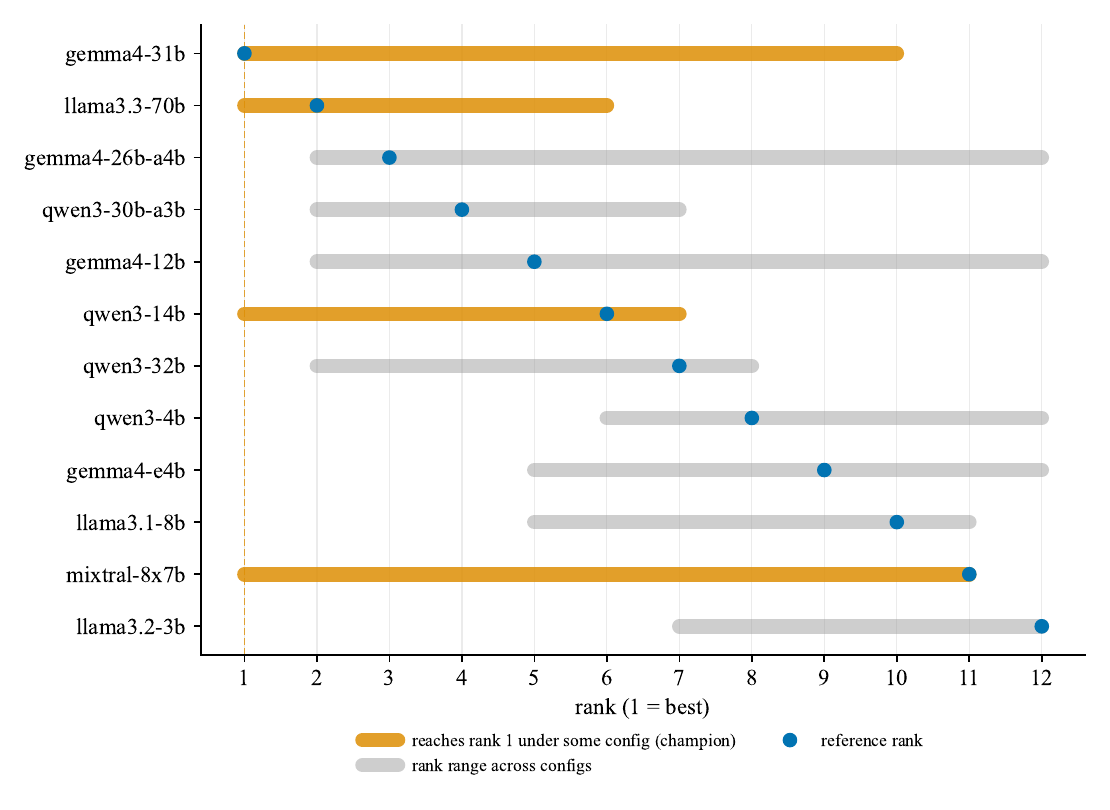}
  \caption{The configuration selects the champion. Models run down the axis in reference-leaderboard order, best at the top, and each bar spans the range of ranks a model attains across the 26 configurations, with a dot at its reference rank. The four models that reach rank one under some configuration are drawn in orange, and the dashed line marks rank one. mixtral-8x7b places 11th of 12 under the reference harness yet reaches first, spanning ranks 1 to 11, and qwen3-14b spans 1 to 7. The winner of the leaderboard is selected by the configuration, not only by the model.}
  \Description{A horizontal chart with one row per model, ordered by reference rank with the best model at the top. Rank runs along the horizontal axis from one to twelve. Each row is a bar spanning the model's minimum to maximum rank across the 26 configurations, with a dot marking the reference rank. Four bars, drawn in orange, reach rank one at the left edge: gemma4-31b, llama3.3-70b, qwen3-14b, and mixtral-8x7b. The mixtral-8x7b bar spans almost the whole width, from rank one to rank eleven.}
  \label{fig:rank}
\end{figure}

If the ordering rests on fragile items, the identity of the best model should itself depend on the configuration. Figure~\ref{fig:rank} shows each model's rank range across the grid. Four distinct models reach rank one: gemma4-31b tops 16 configurations, qwen3-14b tops 8, and llama3.3-70b and mixtral-8x7b top one each. Eight of the 12 reach the top three under some configuration.

A champion count is only worth the margins behind it, so we give them. The two major champions win decisively. gemma4-31b leads its 16 configurations by a median of 181.5 items out of 3{,}679 and never by fewer than 7, and qwen3-14b leads its 8 by a median of 98.5 items and never by fewer than 10. Neither wins any configuration by a margin that a handful of items could erase. The two minor champions are the opposite case: llama3.3-70b tops one configuration by a single item and mixtral-8x7b tops one by three. We count them, because they do attain rank one under a configuration we would defend, and we report their margins so that no reader mistakes a coin flip for a verdict. The claim of this section does not rest on them, and Appendix~\ref{app:margins} tabulates the winner and the margin for all 26 configurations.

What decides between the two major champions is the prompt template. gemma4-31b wins all 6 plain-letter configurations and all 6 parenthesized-letter ones. qwen3-14b wins all 6 digit-labeled ones. In the matched pair of templates that differ in their labels, letter against digit with the same question stem and the same orderings, the winner flips under every one of the 6 option orders. The instruction-phrased template, which also labels its options with letters, is the one generation block the two contest: they divide it 4 to 2, trading first place as the option order changes, with margins there of 7 to 27 items.

The two likelihood configurations do something more drastic than reorder the middle of the ranking. They replace the top of it. Under likelihood scoring the leading pair is llama3.3-70b, second under the reference, and mixtral-8x7b, eleventh, while gemma4-31b, the reference champion, falls to ninth and tenth. Those two sit within 3 items of each other, so which of them is nominally first is a coin flip, and we do not claim otherwise. That the reference leaderboard's eleventh-place model is competitive for first under a scoring rule that ships as the default in common evaluation code \citep{eval-harness} is not a coin flip, and it is the point.

mixtral-8x7b sits 11th of 12 on the reference leaderboard and first under one configuration, with a rank range covering 1 to 11. qwen3-14b ranges from 1 to 7 and gemma4-12b from 2 to 12. The statistic we report is the size of the champion set, not the distance ranks travel, because the two carry different force. A dispersion number says the ranking is noisy. A champion count says that models an evaluator did nothing wrong to select can each be crowned.

\subsection{Compression concentrates the fragility it cannot see}
\label{sec:compression}
Benchmark-compression methods keep the items that best separate strong models from weak ones, which is item discrimination in the item-response-theory sense \citep{rodriguez2021evaluation}, and discard the rest. If those are also the fragile items, compression makes the harness problem worse while appearing to make evaluation cheaper. They are the same items. Discrimination and fragility correlate at 0.28, with a 95 percent confidence interval of 0.25 to 0.30, which is a modest association but a tightly estimated one. The 100 most discriminative items have a mean fragility of 0.96 against an overall mean of 0.85.

The consequence shows up in the champion set, and it has to be stated with care. Discrimination estimated over 12 models takes only 720 distinct values across the 3{,}679 items, so the ranking of items by discrimination is heavily tied. At the rank-100 cutoff just 36 items lie strictly above the threshold while 102 sit exactly on it, which leaves 64 of the 100 slots in a top-100 subset to be filled by how the tie is broken rather than by the data. A single top-100 subset is therefore not well defined. We do not report one draw from it. We marginalize the champion count over 200 seeded random tie-breaks of the items on the cutoff, which makes the statistic a deterministic function of the seed and stable across seeds. Appendix~\ref{app:tiebreak} gives the full distribution over tie-breaks and the sensitivity of its mean to the seed.

Two effects separate in the data. Cutting the benchmark to 100 items at all raises the champion count from 4 on the full set to 5.1 for a random subset of that size, so compression by itself buys volatility. Selecting those 100 items by discrimination raises it further, to a mean of 5.7 with a range of 4 to 7 across tie-breaks. The second effect is the one attributable to the selection criterion, and it is modest: about half a champion beyond what the smaller sample already costs. We report it as such and do not lean on it. The firmer evidence for this claim is the pair of quantities that no tie-break can move, namely the correlation between discrimination and fragility and the fragility gap between the most discriminative items and the rest. Compression does not remove harness sensitivity. It keeps the items that carry it, and it makes the leaderboard built on them more configuration-dependent than the benchmark it replaces.

\subsection{The scoring axis is load-bearing}
\begin{figure}[ht]
  \centering
  \includegraphics[width=0.85\linewidth]{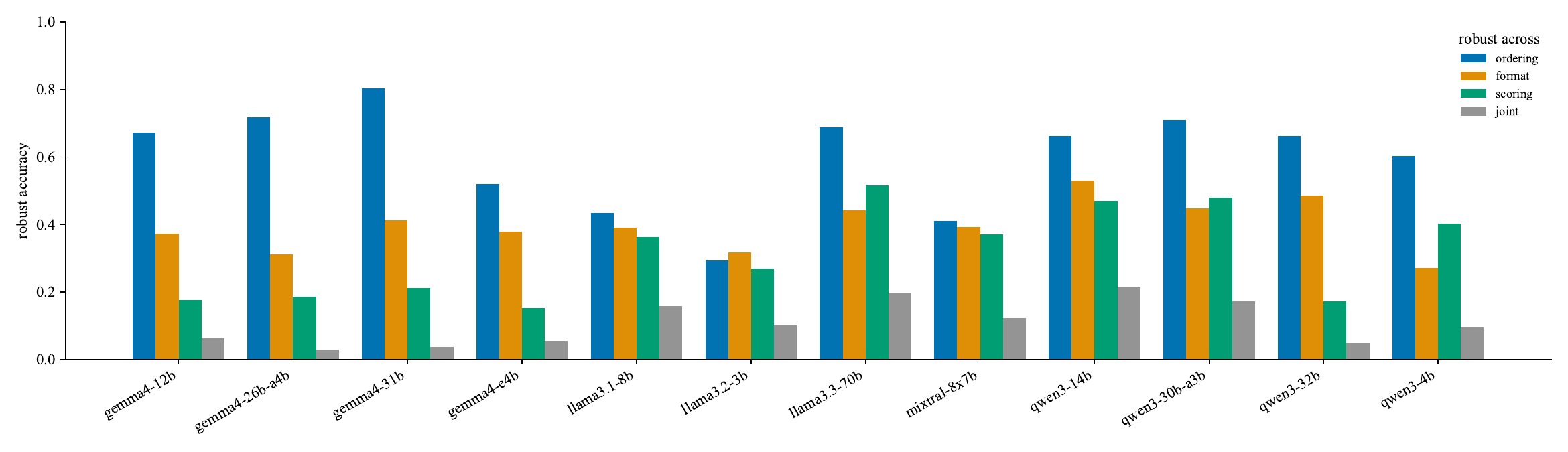}
  \caption{The scoring choice does the damage. For each model, four bars give the accuracy that survives varying one harness axis with the other axes held at the reference: option order, prompt format, scoring method, and the joint grid. Averaged across the 12 models, shuffling the options preserves 0.60 of credited accuracy, prompt format preserves 0.40, and the generation-versus-likelihood choice preserves only 0.31. Varying all three together leaves 0.11. Scoring is the most destructive single axis for 9 of the 12 models, and the joint bar is the smallest for all 12. The axis that evaluation protocols most often fix is the one that matters least.}
  \Description{A grouped bar chart with one group of four bars per model: accuracy robust to option ordering, to prompt format, to scoring method, and to the joint grid. The joint bar is the smallest in every group, and for nine of the twelve models the scoring bar is the lowest of the three single-axis bars.}
  \label{fig:axis}
\end{figure}

The three axes do not contribute equally. For each model we compute axis-robust accuracy, which counts only the items answered correctly under all variants of one axis while the other axes are held at the reference. Table~\ref{tab:axis} gives the values and Figure~\ref{fig:axis} summarizes them.

\begin{table}
  \caption{Axis-robust accuracy: the accuracy that survives varying one axis with the others held at the reference. Joint varies all three at once. Lower is worse, since it means the axis destroys more credited answers. Scoring is the most destructive single axis for 9 of the 12 models.}
  \label{tab:axis}
  \begin{tabular}{@{}lcccc@{}}
    \toprule
    Model & Ordering & Format & Scoring & Joint \\
    \midrule
    gemma4-12b     & 0.673 & 0.373 & 0.177 & 0.062 \\
    gemma4-26b-a4b & 0.718 & 0.310 & 0.186 & 0.029 \\
    gemma4-31b     & 0.803 & 0.412 & 0.213 & 0.037 \\
    gemma4-e4b     & 0.520 & 0.379 & 0.152 & 0.055 \\
    llama3.1-8b    & 0.434 & 0.391 & 0.363 & 0.158 \\
    llama3.2-3b    & 0.294 & 0.316 & 0.269 & 0.100 \\
    llama3.3-70b   & 0.689 & 0.442 & 0.516 & 0.196 \\
    mixtral-8x7b   & 0.411 & 0.392 & 0.370 & 0.122 \\
    qwen3-14b      & 0.662 & 0.529 & 0.470 & 0.215 \\
    qwen3-30b-a3b  & 0.710 & 0.449 & 0.480 & 0.172 \\
    qwen3-32b      & 0.662 & 0.486 & 0.172 & 0.048 \\
    qwen3-4b       & 0.602 & 0.271 & 0.403 & 0.095 \\
    \midrule
    Mean           & 0.598 & 0.396 & 0.314 & 0.108 \\
    \bottomrule
  \end{tabular}
\end{table}

Averaged across models, option order preserves 0.60 of accuracy, prompt format preserves 0.40, and the scoring choice preserves 0.31. Varying all three together leaves 0.11. The scoring choice, not the option order and not the prompt wording, is the load-bearing axis: it is the single axis whose removal from the grid would recover the most credited answers, and it is the axis most often left undocumented. That generation and likelihood diverge so sharply is consistent with the surface-form competition that likelihood scoring introduces \citep{holtzman2021surface}. Protocols that fix the option order and let the evaluator pick the scoring method have controlled the smaller effect and left the larger one free.

The three axes do not span the same number of configurations, and the imbalance runs against us rather than for us. Option order spans 6 configurations, prompt format spans 4, and scoring spans only 3. An axis with more configurations has more chances to flip a given item, so it should mechanically destroy more credited answers. Scoring has the fewest chances of the three and destroys the most, which makes the ranking of the axes a conservative reading of the data rather than an artifact of how the grid was cut.

\subsection{What follows for reporting}
Four changes follow directly from the measurements. A score should be reported as a band with its robust accuracy attached, since the band is what the data supports and the point is one draw from it. A comparison between two models should be made on their jointly robust items, because that is where a difference reflects the models rather than the harness. The harness should be disclosed in full, and the scoring method first, since it is the axis that moves the most. A compressed benchmark should be checked for configuration stability before it is trusted, since the property that makes a subset informative is the property that makes it fragile.

\section{Limitations}

\textbf{Roster scope.} The results cover 12 instruction-tuned models from 4 families between 3 and 70 billion parameters. Two further Mistral models were dropped for an engine-initialization hang rather than for anything about their answers, which leaves Mistral represented by mixtral-8x7b alone, so family-level claims about Mistral rest on one model. We do not claim that the magnitudes reported here transfer unchanged to frontier proprietary models, which we cannot run under a controlled grid.

\textbf{Format scope.} Every claim is about four-option multiple-choice benchmarks, which are the dominant static-evaluation format but not the only one. We do not test open-ended generation, code, or agentic tasks, and we do not claim that a fragility grid would decompose their leaderboards the same way.

\textbf{One benchmark carries a construction artifact.} TruthfulQA is not natively four-option, and our loader assembles each item as the correct answer followed by three distractors, so the gold answer sits first in the benchmark's own order. The reference configuration uses that order, which makes the answer key uniformly the first option on 679 of the 3{,}679 pooled items. The artifact does not favor the leaderboard we report: 10 of the 12 models score \textit{lower} on TruthfulQA under the reference configuration than under a permuted one, and the model it clearly helps is llama3.2-3b, which places last. Recomputing everything on the other three benchmarks leaves all three claims standing and strengthens two of them, but it moves two secondary numbers, since the champion count falls from 4 to 3 and the minimum Kendall correlation rises from $-0.09$ to $+0.03$ (Appendix~\ref{app:loo}). We therefore claim four defensible champions on the item set we evaluate and do not claim that the count is invariant to the benchmark mix.

\textbf{Two of the four champions win by a margin within noise.} The champion count of 4 is not four equal facts. gemma4-31b and qwen3-14b top 16 and 8 of the 26 configurations by median margins of 181.5 and 98.5 items out of 3{,}679, and that split is what carries the claim. llama3.3-70b and mixtral-8x7b top one configuration each, by 1 item and by 3 items, and on a three-benchmark item set one of those wins becomes an exact tie broken by sort order (Appendix~\ref{app:loo}). We count them because they attain rank one under a configuration we would defend, and we report every margin in Appendix~\ref{app:margins}. We do not claim that the number 4 is itself robust. We claim that the harness selects the winner, which the 24 configurations split between the top two establish on their own.

\textbf{The grid is a judgment.} The 26 configurations are ones a practitioner uses and would defend, and none is adversarial, which is what gives the champion count its force. A reader who considers likelihood scoring illegitimate for chat models would drop 2 of the 26 configurations and see a smaller effect. We state the grid in full so that this line can be redrawn, and we do not claim that our boundary is the only defensible one.

\textbf{The robust-item tie is a small-sample statement, and the interval is the weaker half of it.} Each pair has only 72 to 336 jointly robust items, and on those items the observed disagreement is zero for five pairs and a single item for four more. We report those counts rather than lean on the bootstrap interval, because for the nine pairs with no disagreement in one direction the interval's lower bound is pinned at zero by construction and could not have excluded zero whatever the data said. The interval is therefore consistent with the claim but is not independent evidence for it. A larger jointly robust set, which a wider grid would not produce and a narrower one would, could still separate some of these pairs. We claim that the ordering is unidentified on the evidence available, and we do not claim that the models are provably equal on those items.

\textbf{Decoding.} Decoding is greedy and single-seed by design, since that is how static leaderboards are scored, so sampling variance is excluded from the grid rather than measured. We do not claim that harness variance is larger than sampling variance, only that harness variance alone suffices to manufacture the ordering.

\textbf{The discriminative subset is not uniquely defined.} Discrimination estimated over 12 models takes only 720 distinct values across 3{,}679 items, and 102 items tie exactly at the rank-100 cutoff, so 64 of the 100 slots in a top-100 subset are set by the tie-break and not by the data. The champion count for that subset ranges from 4 to 7 across seeded tie-breaks, with a mean of 5.7. We report the distribution rather than a single draw, and we do not claim a precise champion count for a compressed benchmark. A larger model pool would estimate discrimination more finely and could sharpen this. The correlation between discrimination and fragility and the fragility gap between the most discriminative items and the rest do not depend on the tie-break, and the compression claim rests on those.

\textbf{The scoring axis is two likelihood configurations.} The scoring axis holds the reference generation configuration against two likelihood cloze stems, which is the dichotomy the field actually faces, since likelihood scoring is the default in common evaluation code. It does mean that a single family of scorers produces much of the effect we attribute to the axis, including the lowest single-configuration score in the roster. We do not claim that every likelihood implementation would behave this way.

\section{Future Work}
\label{sec:future}
The margin between a model's top two options is recorded for every item in our release and offers a route to a per-item flip predictor, which would let an evaluator flag fragile items before running any sweep and turn the grid into a cheap screening step. Separating wrong answers from unparseable ones would split the scoring axis into a competence term and a compliance term and attribute the ranking movement to each, which the present decomposition treats as one axis. The grid holds few-shot prompting and chain-of-thought fixed, and both add harness choices that a wider grid would cover. The same robust-versus-fragile split applies to preference and arena leaderboards, where the analog of a configuration is the judge or the aggregation rule. The champion count states the size of the problem and invites a matching remedy, since an aggregation rule that returns a configuration-robust winner would be evaluated naturally against exactly this statistic.

\section{Conclusion}
A multiple-choice leaderboard is read as a measurement of models. It is a measurement of models under a harness, and the harness is doing more of the work than the number admits. Varying it across 26 configurations that any evaluator would defend, we find that neighboring models are tied on every item they answer stably, and that all 10 adjacent gaps that exist at all are carried by the items the two models are least sure about, at 95.7 percent of the mean gap. Four of 12 models can each be crowned by choosing a configuration, and in the matched pair of templates that differ in their labels, swapping letters for digits flips which of the two leading models wins under every option order. The items that benchmark compression retains are the fragile ones, so the cheap leaderboard is the volatile one, and the scoring method rather than the option order is the axis that does the damage. None of this says the models are equal. It says the order is not in the data, and a leaderboard that prints one number from one harness should print the band, the robust accuracy, and the harness alongside it.

\section*{Code and Data Availability}
The full release is at \url{https://github.com/NikolaTesla-007/fragility-grid}. It holds the per-item correctness records, which are 48 files with 26 configuration bits for every item and model, the configuration legend, the analysis script with its unit tests, and the code that renders the four figures. Every number and figure in this paper regenerates from the records on a CPU in about ten seconds, with no GPU and no model download, and two runs give byte-identical output. The records embed the benchmark question texts under the source benchmarks' own licenses, and the code is released under the MIT license.

\bibliographystyle{plainnat}
\bibliography{references}

\begin{thebibliography}{30}
\providecommand{\natexlab}[1]{#1}
\providecommand{\url}[1]{\texttt{#1}}
\expandafter\ifx\csname urlstyle\endcsname\relax
  \providecommand{\doi}[1]{doi: #1}\else
  \providecommand{\doi}{doi: \begingroup \urlstyle{rm}\Url}\fi

\bibitem[Alzahrani et~al.(2024)Alzahrani, Alyahya, Alnumay, AlRashed, Alsubaie,
  Almushayqih, Mirza, Alotaibi, Al-Twairesh, Alowisheq, Bari, and
  Khan]{alzahrani2024targets}
Norah Alzahrani, Hisham Alyahya, Yazeed Alnumay, Sultan AlRashed, Shaykhah
  Alsubaie, Yousef Almushayqih, Faisal Mirza, Nouf Alotaibi, Nora Al-Twairesh,
  Areeb Alowisheq, M~Saiful Bari, and Haidar Khan.
\newblock When benchmarks are targets: Revealing the sensitivity of large
  language model leaderboards.
\newblock In Lun-Wei Ku, Andre Martins, and Vivek Srikumar, editors,
  \emph{Proceedings of the 62nd Annual Meeting of the Association for
  Computational Linguistics (Volume 1: Long Papers)}, pages 13787--13805,
  Bangkok, Thailand, August 2024. Association for Computational Linguistics.
\newblock \doi{10.18653/v1/2024.acl-long.744}.
\newblock URL \url{https://aclanthology.org/2024.acl-long.744/}.

\bibitem[Bowman and Dahl(2021)]{bowman2021benchmarking}
Samuel~R. Bowman and George Dahl.
\newblock What will it take to fix benchmarking in natural language
  understanding?
\newblock In Kristina Toutanova, Anna Rumshisky, Luke Zettlemoyer, Dilek
  Hakkani-Tur, Iz~Beltagy, Steven Bethard, Ryan Cotterell, Tanmoy Chakraborty,
  and Yichao Zhou, editors, \emph{Proceedings of the 2021 Conference of the
  North American Chapter of the Association for Computational Linguistics:
  Human Language Technologies}, pages 4843--4855, Online, June 2021.
  Association for Computational Linguistics.
\newblock \doi{10.18653/v1/2021.naacl-main.385}.
\newblock URL \url{https://aclanthology.org/2021.naacl-main.385/}.

\bibitem[Card et~al.(2020)Card, Henderson, Khandelwal, Jia, Mahowald, and
  Jurafsky]{card2020power}
Dallas Card, Peter Henderson, Urvashi Khandelwal, Robin Jia, Kyle Mahowald, and
  Dan Jurafsky.
\newblock With little power comes great responsibility.
\newblock In Bonnie Webber, Trevor Cohn, Yulan He, and Yang Liu, editors,
  \emph{Proceedings of the 2020 Conference on Empirical Methods in Natural
  Language Processing (EMNLP)}, pages 9263--9274, Online, November 2020.
  Association for Computational Linguistics.
\newblock \doi{10.18653/v1/2020.emnlp-main.745}.
\newblock URL \url{https://aclanthology.org/2020.emnlp-main.745/}.

\bibitem[Chiang et~al.(2024)Chiang, Zheng, Sheng, Angelopoulos, Li, Li, Zhu,
  Zhang, Jordan, Gonzalez, and Stoica]{chiang2024arena}
Wei-Lin Chiang, Lianmin Zheng, Ying Sheng, Anastasios~N. Angelopoulos, Tianle
  Li, Dacheng Li, Banghua Zhu, Hao Zhang, Michael~I. Jordan, Joseph~E.
  Gonzalez, and Ion Stoica.
\newblock Chatbot arena: an open platform for evaluating llms by human
  preference.
\newblock In \emph{Proceedings of the 41st International Conference on Machine
  Learning}, ICML'24. JMLR.org, 2024.

\bibitem[Clark et~al.(2018)Clark, Cowhey, Etzioni, Khot, Sabharwal, Schoenick,
  and Tafjord]{clark2018arc}
Peter Clark, Isaac Cowhey, Oren Etzioni, Tushar Khot, Ashish Sabharwal, Carissa
  Schoenick, and Oyvind Tafjord.
\newblock Think you have solved question answering? try arc, the ai2 reasoning
  challenge, 2018.
\newblock URL \url{https://arxiv.org/abs/1803.05457}.

\bibitem[Dror et~al.(2018)Dror, Baumer, Shlomov, and
  Reichart]{dror2018hitchhiker}
Rotem Dror, Gili Baumer, Segev Shlomov, and Roi Reichart.
\newblock The hitchhiker{'}s guide to testing statistical significance in
  natural language processing.
\newblock In Iryna Gurevych and Yusuke Miyao, editors, \emph{Proceedings of the
  56th Annual Meeting of the Association for Computational Linguistics (Volume
  1: Long Papers)}, pages 1383--1392, Melbourne, Australia, July 2018.
  Association for Computational Linguistics.
\newblock \doi{10.18653/v1/P18-1128}.
\newblock URL \url{https://aclanthology.org/P18-1128/}.

\bibitem[Efron and Tibshirani(1994)]{efron1994bootstrap}
Bradley Efron and Robert~J. Tibshirani.
\newblock \emph{An Introduction to the Bootstrap}.
\newblock Chapman and Hall/CRC, 1994.

\bibitem[Embretson and Reise(2000)]{embretson2000irt}
Susan~E. Embretson and Steven~P. Reise.
\newblock \emph{Item Response Theory for Psychologists}.
\newblock Lawrence Erlbaum Associates, 2000.

\bibitem[Gao et~al.(2023)Gao, Tow, Abbasi, Biderman, Black, DiPofi, Foster,
  Golding, Hsu, Le~Noac'h, Li, McDonell, Muennighoff, Ociepa, Phang, Reynolds,
  Schoelkopf, Skowron, Sutawika, Tang, Thite, Wang, Wang, and
  Zou]{eval-harness}
Leo Gao, Jonathan Tow, Baber Abbasi, Stella Biderman, Sid Black, Anthony
  DiPofi, Charles Foster, Laurence Golding, Jeffrey Hsu, Alain Le~Noac'h,
  Haonan Li, Kyle McDonell, Niklas Muennighoff, Chris Ociepa, Jason Phang,
  Laria Reynolds, Hailey Schoelkopf, Aviya Skowron, Lintang Sutawika, Eric
  Tang, Anish Thite, Ben Wang, Kevin Wang, and Andy Zou.
\newblock A framework for few-shot language model evaluation, 12 2023.
\newblock URL \url{https://zenodo.org/records/10256836}.

\bibitem[Grattafiori et~al.(2024)Grattafiori, Dubey, Jauhri, Pandey, Kadian,
  Al-Dahle, Letman, Mathur, Schelten, Vaughan, Yang, Fan, Goyal, Hartshorn,
  Yang, Mitra, Sravankumar, Korenev, Hinsvark, Rao, Zhang, Rodriguez,
  Gregerson, Spataru, Roziere, Biron, Tang, Chern, Caucheteux, Nayak, Bi,
  Marra, McConnell, Keller, Touret, Wu, Wong, Ferrer, Nikolaidis, Allonsius,
  Song, Pintz, Livshits, Wyatt, Esiobu, Choudhary, Mahajan, Garcia-Olano,
  Perino, Hupkes, Lakomkin, AlBadawy, Lobanova, Dinan, Smith, Radenovic,
  Guzmán, Zhang, Synnaeve, Lee, Anderson, Thattai, Nail, Mialon, Pang,
  Cucurell, Nguyen, Korevaar, Xu, Touvron, Zarov, Ibarra, Kloumann, Misra,
  Evtimov, Zhang, Copet, Lee, Geffert, Vranes, Park, Mahadeokar, Shah, van~der
  Linde, Billock, Hong, Lee, Fu, Chi, Huang, Liu, Wang, Yu, Bitton, Spisak,
  Park, Rocca, Johnstun, Saxe, Jia, Alwala, Prasad, Upasani, Plawiak, Li,
  Heafield, Stone, El-Arini, Iyer, Malik, Chiu, Bhalla, Lakhotia,
  Rantala-Yeary, van~der Maaten, Chen, Tan, Jenkins, Martin, Madaan, Malo,
  Blecher, Landzaat, de~Oliveira, Muzzi, Pasupuleti, Singh, Paluri, Kardas,
  Tsimpoukelli, Oldham, Rita, Pavlova, Kambadur, Lewis, Si, Singh, Hassan,
  Goyal, Torabi, Bashlykov, Bogoychev, Chatterji, Zhang, Duchenne, Çelebi,
  Alrassy, Zhang, Li, Vasic, Weng, Bhargava, Dubal, Krishnan, Koura, Xu, He,
  Dong, Srinivasan, Ganapathy, Calderer, Cabral, Stojnic, Raileanu, Maheswari,
  Girdhar, Patel, Sauvestre, Polidoro, Sumbaly, Taylor, Silva, Hou, Wang,
  Hosseini, Chennabasappa, Singh, Bell, Kim, Edunov, Nie, Narang, Raparthy,
  Shen, Wan, Bhosale, Zhang, Vandenhende, Batra, Whitman, Sootla, Collot,
  Gururangan, Borodinsky, Herman, Fowler, Sheasha, Georgiou, Scialom,
  Speckbacher, Mihaylov, Xiao, Karn, Goswami, Gupta, Ramanathan, Kerkez,
  Gonguet, Do, Vogeti, Albiero, Petrovic, Chu, Xiong, Fu, Meers, Martinet,
  Wang, Wang, Tan, Xia, Xie, Jia, Wang, Goldschlag, Gaur, Babaei, Wen, Song,
  Zhang, Li, Mao, Coudert, Yan, Chen, Papakipos, Singh, Srivastava, Jain,
  Kelsey, Shajnfeld, Gangidi, Victoria, Goldstand, Menon, Sharma, Boesenberg,
  Baevski, Feinstein, Kallet, Sangani, Teo, Yunus, Lupu, Alvarado, Caples, Gu,
  Ho, Poulton, Ryan, Ramchandani, Dong, Franco, Goyal, Saraf, Chowdhury,
  Gabriel, Bharambe, Eisenman, Yazdan, James, Maurer, Leonhardi, Huang, Loyd,
  Paola, Paranjape, Liu, Wu, Ni, Hancock, Wasti, Spence, Stojkovic, Gamido,
  Montalvo, Parker, Burton, Mejia, Liu, Wang, Kim, Zhou, Hu, Chu, Cai, Tindal,
  Feichtenhofer, Gao, Civin, Beaty, Kreymer, Li, Adkins, Xu, Testuggine, David,
  Parikh, Liskovich, Foss, Wang, Le, Holland, Dowling, Jamil, Montgomery,
  Presani, Hahn, Wood, Le, Brinkman, Arcaute, Dunbar, Smothers, Sun, Kreuk,
  Tian, Kokkinos, Ozgenel, Caggioni, Kanayet, Seide, Florez, Schwarz, Badeer,
  Swee, Halpern, Herman, Sizov, Guangyi, Zhang, Lakshminarayanan, Inan,
  Shojanazeri, Zou, Wang, Zha, Habeeb, Rudolph, Suk, Aspegren, Goldman, Zhan,
  Damlaj, Molybog, Tufanov, Leontiadis, Veliche, Gat, Weissman, Geboski, Kohli,
  Lam, Asher, Gaya, Marcus, Tang, Chan, Zhen, Reizenstein, Teboul, Zhong, Jin,
  Yang, Cummings, Carvill, Shepard, McPhie, Torres, Ginsburg, Wang, Wu, U,
  Saxena, Khandelwal, Zand, Matosich, Veeraraghavan, Michelena, Li, Jagadeesh,
  Huang, Chawla, Huang, Chen, Garg, A, Silva, Bell, Zhang, Guo, Yu, Moshkovich,
  Wehrstedt, Khabsa, Avalani, Bhatt, Mankus, Hasson, Lennie, Reso, Groshev,
  Naumov, Lathi, Keneally, Liu, Seltzer, Valko, Restrepo, Patel, Vyatskov,
  Samvelyan, Clark, Macey, Wang, Hermoso, Metanat, Rastegari, Bansal,
  Santhanam, Parks, White, Bawa, Singhal, Egebo, Usunier, Mehta, Laptev, Dong,
  Cheng, Chernoguz, Hart, Salpekar, Kalinli, Kent, Parekh, Saab, Balaji,
  Rittner, Bontrager, Roux, Dollar, Zvyagina, Ratanchandani, Yuvraj, Liang,
  Alao, Rodriguez, Ayub, Murthy, Nayani, Mitra, Parthasarathy, Li, Hogan,
  Battey, Wang, Howes, Rinott, Mehta, Siby, Bondu, Datta, Chugh, Hunt, Dhillon,
  Sidorov, Pan, Mahajan, Verma, Yamamoto, Ramaswamy, Lindsay, Lindsay, Feng,
  Lin, Zha, Patil, Shankar, Zhang, Zhang, Wang, Agarwal, Sajuyigbe, Chintala,
  Max, Chen, Kehoe, Satterfield, Govindaprasad, Gupta, Deng, Cho, Virk,
  Subramanian, Choudhury, Goldman, Remez, Glaser, Best, Koehler, Robinson, Li,
  Zhang, Matthews, Chou, Shaked, Vontimitta, Ajayi, Montanez, Mohan, Kumar,
  Mangla, Ionescu, Poenaru, Mihailescu, Ivanov, Li, Wang, Jiang, Bouaziz,
  Constable, Tang, Wu, Wang, Wu, Gao, Kleinman, Chen, Hu, Jia, Qi, Li, Zhang,
  Zhang, Adi, Nam, Yu, Wang, Zhao, Hao, Qian, Li, He, Rait, DeVito, Rosnbrick,
  Wen, Yang, Zhao, and Ma]{dubey2024llama3}
Aaron Grattafiori, Abhimanyu Dubey, Abhinav Jauhri, Abhinav Pandey, Abhishek
  Kadian, Ahmad Al-Dahle, Aiesha Letman, Akhil Mathur, Alan Schelten, Alex
  Vaughan, Amy Yang, Angela Fan, Anirudh Goyal, Anthony Hartshorn, Aobo Yang,
  Archi Mitra, Archie Sravankumar, Artem Korenev, Arthur Hinsvark, Arun Rao,
  Aston Zhang, Aurelien Rodriguez, Austen Gregerson, Ava Spataru, Baptiste
  Roziere, Bethany Biron, Binh Tang, Bobbie Chern, Charlotte Caucheteux, Chaya
  Nayak, Chloe Bi, Chris Marra, Chris McConnell, Christian Keller, Christophe
  Touret, Chunyang Wu, Corinne Wong, Cristian~Canton Ferrer, Cyrus Nikolaidis,
  Damien Allonsius, Daniel Song, Danielle Pintz, Danny Livshits, Danny Wyatt,
  David Esiobu, Dhruv Choudhary, Dhruv Mahajan, Diego Garcia-Olano, Diego
  Perino, Dieuwke Hupkes, Egor Lakomkin, Ehab AlBadawy, Elina Lobanova, Emily
  Dinan, Eric~Michael Smith, Filip Radenovic, Francisco Guzmán, Frank Zhang,
  Gabriel Synnaeve, Gabrielle Lee, Georgia~Lewis Anderson, Govind Thattai,
  Graeme Nail, Gregoire Mialon, Guan Pang, Guillem Cucurell, Hailey Nguyen,
  Hannah Korevaar, Hu~Xu, Hugo Touvron, Iliyan Zarov, Imanol~Arrieta Ibarra,
  Isabel Kloumann, Ishan Misra, Ivan Evtimov, Jack Zhang, Jade Copet, Jaewon
  Lee, Jan Geffert, Jana Vranes, Jason Park, Jay Mahadeokar, Jeet Shah, Jelmer
  van~der Linde, Jennifer Billock, Jenny Hong, Jenya Lee, Jeremy Fu, Jianfeng
  Chi, Jianyu Huang, Jiawen Liu, Jie Wang, Jiecao Yu, Joanna Bitton, Joe
  Spisak, Jongsoo Park, Joseph Rocca, Joshua Johnstun, Joshua Saxe, Junteng
  Jia, Kalyan~Vasuden Alwala, Karthik Prasad, Kartikeya Upasani, Kate Plawiak,
  Ke~Li, Kenneth Heafield, Kevin Stone, Khalid El-Arini, Krithika Iyer, Kshitiz
  Malik, Kuenley Chiu, Kunal Bhalla, Kushal Lakhotia, Lauren Rantala-Yeary,
  Laurens van~der Maaten, Lawrence Chen, Liang Tan, Liz Jenkins, Louis Martin,
  Lovish Madaan, Lubo Malo, Lukas Blecher, Lukas Landzaat, Luke de~Oliveira,
  Madeline Muzzi, Mahesh Pasupuleti, Mannat Singh, Manohar Paluri, Marcin
  Kardas, Maria Tsimpoukelli, Mathew Oldham, Mathieu Rita, Maya Pavlova,
  Melanie Kambadur, Mike Lewis, Min Si, Mitesh~Kumar Singh, Mona Hassan, Naman
  Goyal, Narjes Torabi, Nikolay Bashlykov, Nikolay Bogoychev, Niladri
  Chatterji, Ning Zhang, Olivier Duchenne, Onur Çelebi, Patrick Alrassy,
  Pengchuan Zhang, Pengwei Li, Petar Vasic, Peter Weng, Prajjwal Bhargava,
  Pratik Dubal, Praveen Krishnan, Punit~Singh Koura, Puxin Xu, Qing He,
  Qingxiao Dong, Ragavan Srinivasan, Raj Ganapathy, Ramon Calderer,
  Ricardo~Silveira Cabral, Robert Stojnic, Roberta Raileanu, Rohan Maheswari,
  Rohit Girdhar, Rohit Patel, Romain Sauvestre, Ronnie Polidoro, Roshan
  Sumbaly, Ross Taylor, Ruan Silva, Rui Hou, Rui Wang, Saghar Hosseini, Sahana
  Chennabasappa, Sanjay Singh, Sean Bell, Seohyun~Sonia Kim, Sergey Edunov,
  Shaoliang Nie, Sharan Narang, Sharath Raparthy, Sheng Shen, Shengye Wan,
  Shruti Bhosale, Shun Zhang, Simon Vandenhende, Soumya Batra, Spencer Whitman,
  Sten Sootla, Stephane Collot, Suchin Gururangan, Sydney Borodinsky, Tamar
  Herman, Tara Fowler, Tarek Sheasha, Thomas Georgiou, Thomas Scialom, Tobias
  Speckbacher, Todor Mihaylov, Tong Xiao, Ujjwal Karn, Vedanuj Goswami, Vibhor
  Gupta, Vignesh Ramanathan, Viktor Kerkez, Vincent Gonguet, Virginie Do, Vish
  Vogeti, Vítor Albiero, Vladan Petrovic, Weiwei Chu, Wenhan Xiong, Wenyin Fu,
  Whitney Meers, Xavier Martinet, Xiaodong Wang, Xiaofang Wang, Xiaoqing~Ellen
  Tan, Xide Xia, Xinfeng Xie, Xuchao Jia, Xuewei Wang, Yaelle Goldschlag,
  Yashesh Gaur, Yasmine Babaei, Yi~Wen, Yiwen Song, Yuchen Zhang, Yue Li,
  Yuning Mao, Zacharie~Delpierre Coudert, Zheng Yan, Zhengxing Chen, Zoe
  Papakipos, Aaditya Singh, Aayushi Srivastava, Abha Jain, Adam Kelsey, Adam
  Shajnfeld, Adithya Gangidi, Adolfo Victoria, Ahuva Goldstand, Ajay Menon,
  Ajay Sharma, Alex Boesenberg, Alexei Baevski, Allie Feinstein, Amanda Kallet,
  Amit Sangani, Amos Teo, Anam Yunus, Andrei Lupu, Andres Alvarado, Andrew
  Caples, Andrew Gu, Andrew Ho, Andrew Poulton, Andrew Ryan, Ankit Ramchandani,
  Annie Dong, Annie Franco, Anuj Goyal, Aparajita Saraf, Arkabandhu Chowdhury,
  Ashley Gabriel, Ashwin Bharambe, Assaf Eisenman, Azadeh Yazdan, Beau James,
  Ben Maurer, Benjamin Leonhardi, Bernie Huang, Beth Loyd, Beto~De Paola,
  Bhargavi Paranjape, Bing Liu, Bo~Wu, Boyu Ni, Braden Hancock, Bram Wasti,
  Brandon Spence, Brani Stojkovic, Brian Gamido, Britt Montalvo, Carl Parker,
  Carly Burton, Catalina Mejia, Ce~Liu, Changhan Wang, Changkyu Kim, Chao Zhou,
  Chester Hu, Ching-Hsiang Chu, Chris Cai, Chris Tindal, Christoph
  Feichtenhofer, Cynthia Gao, Damon Civin, Dana Beaty, Daniel Kreymer, Daniel
  Li, David Adkins, David Xu, Davide Testuggine, Delia David, Devi Parikh,
  Diana Liskovich, Didem Foss, Dingkang Wang, Duc Le, Dustin Holland, Edward
  Dowling, Eissa Jamil, Elaine Montgomery, Eleonora Presani, Emily Hahn, Emily
  Wood, Eric-Tuan Le, Erik Brinkman, Esteban Arcaute, Evan Dunbar, Evan
  Smothers, Fei Sun, Felix Kreuk, Feng Tian, Filippos Kokkinos, Firat Ozgenel,
  Francesco Caggioni, Frank Kanayet, Frank Seide, Gabriela~Medina Florez,
  Gabriella Schwarz, Gada Badeer, Georgia Swee, Gil Halpern, Grant Herman,
  Grigory Sizov, Guangyi, Zhang, Guna Lakshminarayanan, Hakan Inan, Hamid
  Shojanazeri, Han Zou, Hannah Wang, Hanwen Zha, Haroun Habeeb, Harrison
  Rudolph, Helen Suk, Henry Aspegren, Hunter Goldman, Hongyuan Zhan, Ibrahim
  Damlaj, Igor Molybog, Igor Tufanov, Ilias Leontiadis, Irina-Elena Veliche,
  Itai Gat, Jake Weissman, James Geboski, James Kohli, Janice Lam, Japhet
  Asher, Jean-Baptiste Gaya, Jeff Marcus, Jeff Tang, Jennifer Chan, Jenny Zhen,
  Jeremy Reizenstein, Jeremy Teboul, Jessica Zhong, Jian Jin, Jingyi Yang, Joe
  Cummings, Jon Carvill, Jon Shepard, Jonathan McPhie, Jonathan Torres, Josh
  Ginsburg, Junjie Wang, Kai Wu, Kam~Hou U, Karan Saxena, Kartikay Khandelwal,
  Katayoun Zand, Kathy Matosich, Kaushik Veeraraghavan, Kelly Michelena, Keqian
  Li, Kiran Jagadeesh, Kun Huang, Kunal Chawla, Kyle Huang, Lailin Chen,
  Lakshya Garg, Lavender A, Leandro Silva, Lee Bell, Lei Zhang, Liangpeng Guo,
  Licheng Yu, Liron Moshkovich, Luca Wehrstedt, Madian Khabsa, Manav Avalani,
  Manish Bhatt, Martynas Mankus, Matan Hasson, Matthew Lennie, Matthias Reso,
  Maxim Groshev, Maxim Naumov, Maya Lathi, Meghan Keneally, Miao Liu,
  Michael~L. Seltzer, Michal Valko, Michelle Restrepo, Mihir Patel, Mik
  Vyatskov, Mikayel Samvelyan, Mike Clark, Mike Macey, Mike Wang, Miquel~Jubert
  Hermoso, Mo~Metanat, Mohammad Rastegari, Munish Bansal, Nandhini Santhanam,
  Natascha Parks, Natasha White, Navyata Bawa, Nayan Singhal, Nick Egebo,
  Nicolas Usunier, Nikhil Mehta, Nikolay~Pavlovich Laptev, Ning Dong, Norman
  Cheng, Oleg Chernoguz, Olivia Hart, Omkar Salpekar, Ozlem Kalinli, Parkin
  Kent, Parth Parekh, Paul Saab, Pavan Balaji, Pedro Rittner, Philip Bontrager,
  Pierre Roux, Piotr Dollar, Polina Zvyagina, Prashant Ratanchandani, Pritish
  Yuvraj, Qian Liang, Rachad Alao, Rachel Rodriguez, Rafi Ayub, Raghotham
  Murthy, Raghu Nayani, Rahul Mitra, Rangaprabhu Parthasarathy, Raymond Li,
  Rebekkah Hogan, Robin Battey, Rocky Wang, Russ Howes, Ruty Rinott, Sachin
  Mehta, Sachin Siby, Sai~Jayesh Bondu, Samyak Datta, Sara Chugh, Sara Hunt,
  Sargun Dhillon, Sasha Sidorov, Satadru Pan, Saurabh Mahajan, Saurabh Verma,
  Seiji Yamamoto, Sharadh Ramaswamy, Shaun Lindsay, Shaun Lindsay, Sheng Feng,
  Shenghao Lin, Shengxin~Cindy Zha, Shishir Patil, Shiva Shankar, Shuqiang
  Zhang, Shuqiang Zhang, Sinong Wang, Sneha Agarwal, Soji Sajuyigbe, Soumith
  Chintala, Stephanie Max, Stephen Chen, Steve Kehoe, Steve Satterfield,
  Sudarshan Govindaprasad, Sumit Gupta, Summer Deng, Sungmin Cho, Sunny Virk,
  Suraj Subramanian, Sy~Choudhury, Sydney Goldman, Tal Remez, Tamar Glaser,
  Tamara Best, Thilo Koehler, Thomas Robinson, Tianhe Li, Tianjun Zhang, Tim
  Matthews, Timothy Chou, Tzook Shaked, Varun Vontimitta, Victoria Ajayi,
  Victoria Montanez, Vijai Mohan, Vinay~Satish Kumar, Vishal Mangla, Vlad
  Ionescu, Vlad Poenaru, Vlad~Tiberiu Mihailescu, Vladimir Ivanov, Wei Li,
  Wenchen Wang, Wenwen Jiang, Wes Bouaziz, Will Constable, Xiaocheng Tang,
  Xiaojian Wu, Xiaolan Wang, Xilun Wu, Xinbo Gao, Yaniv Kleinman, Yanjun Chen,
  Ye~Hu, Ye~Jia, Ye~Qi, Yenda Li, Yilin Zhang, Ying Zhang, Yossi Adi, Youngjin
  Nam, Yu, Wang, Yu~Zhao, Yuchen Hao, Yundi Qian, Yunlu Li, Yuzi He, Zach Rait,
  Zachary DeVito, Zef Rosnbrick, Zhaoduo Wen, Zhenyu Yang, Zhiwei Zhao, and
  Zhiyu Ma.
\newblock The llama 3 herd of models, 2024.
\newblock URL \url{https://arxiv.org/abs/2407.21783}.

\bibitem[Hendrycks et~al.(2021)Hendrycks, Burns, Basart, Zou, Mazeika, Song,
  and Steinhardt]{hendrycks2021mmlu}
Dan Hendrycks, Collin Burns, Steven Basart, Andy Zou, Mantas Mazeika, Dawn
  Song, and Jacob Steinhardt.
\newblock Measuring massive multitask language understanding, 2021.
\newblock URL \url{https://arxiv.org/abs/2009.03300}.

\bibitem[Holtzman et~al.(2021)Holtzman, West, Shwartz, Choi, and
  Zettlemoyer]{holtzman2021surface}
Ari Holtzman, Peter West, Vered Shwartz, Yejin Choi, and Luke Zettlemoyer.
\newblock Surface form competition: Why the highest probability answer isn{'}t
  always right.
\newblock In Marie-Francine Moens, Xuanjing Huang, Lucia Specia, and Scott
  Wen-tau Yih, editors, \emph{Proceedings of the 2021 Conference on Empirical
  Methods in Natural Language Processing}, pages 7038--7051, Online and Punta
  Cana, Dominican Republic, November 2021. Association for Computational
  Linguistics.
\newblock \doi{10.18653/v1/2021.emnlp-main.564}.
\newblock URL \url{https://aclanthology.org/2021.emnlp-main.564/}.

\bibitem[Jiang et~al.(2024)Jiang, Sablayrolles, Roux, Mensch, Savary, Bamford,
  Chaplot, de~las Casas, Hanna, Bressand, Lengyel, Bour, Lample, Lavaud,
  Saulnier, Lachaux, Stock, Subramanian, Yang, Antoniak, Scao, Gervet, Lavril,
  Wang, Lacroix, and Sayed]{jiang2024mixtral}
Albert~Q. Jiang, Alexandre Sablayrolles, Antoine Roux, Arthur Mensch, Blanche
  Savary, Chris Bamford, Devendra~Singh Chaplot, Diego de~las Casas, Emma~Bou
  Hanna, Florian Bressand, Gianna Lengyel, Guillaume Bour, Guillaume Lample,
  Lélio~Renard Lavaud, Lucile Saulnier, Marie-Anne Lachaux, Pierre Stock,
  Sandeep Subramanian, Sophia Yang, Szymon Antoniak, Teven~Le Scao, Théophile
  Gervet, Thibaut Lavril, Thomas Wang, Timothée Lacroix, and William~El Sayed.
\newblock Mixtral of experts, 2024.
\newblock URL \url{https://arxiv.org/abs/2401.04088}.

\bibitem[Kwon et~al.(2023)Kwon, Li, Zhuang, Sheng, Zheng, Yu, Gonzalez, Zhang,
  and Stoica]{kwon2023vllm}
Woosuk Kwon, Zhuohan Li, Siyuan Zhuang, Ying Sheng, Lianmin Zheng, Cody~Hao Yu,
  Joseph Gonzalez, Hao Zhang, and Ion Stoica.
\newblock Efficient memory management for large language model serving with
  pagedattention.
\newblock In \emph{Proceedings of the 29th Symposium on Operating Systems
  Principles}, SOSP '23, page 611–626, New York, NY, USA, 2023. Association
  for Computing Machinery.
\newblock ISBN 9798400702297.
\newblock \doi{10.1145/3600006.3613165}.
\newblock URL \url{https://doi.org/10.1145/3600006.3613165}.

\bibitem[Liang et~al.(2023)Liang, Bommasani, Lee, Tsipras, Soylu, Yasunaga,
  Zhang, Narayanan, Wu, Kumar, Newman, Yuan, Yan, Zhang, Cosgrove, Manning, Re,
  Acosta-Navas, Hudson, Zelikman, Durmus, Ladhak, Rong, Ren, Yao, WANG,
  Santhanam, Orr, Zheng, Yuksekgonul, Suzgun, Kim, Guha, Chatterji, Khattab,
  Henderson, Huang, Chi, Xie, Santurkar, Ganguli, Hashimoto, Icard, Zhang,
  Chaudhary, Wang, Li, Mai, Zhang, and Koreeda]{liang2023helm}
Percy Liang, Rishi Bommasani, Tony Lee, Dimitris Tsipras, Dilara Soylu,
  Michihiro Yasunaga, Yian Zhang, Deepak Narayanan, Yuhuai Wu, Ananya Kumar,
  Benjamin Newman, Binhang Yuan, Bobby Yan, Ce~Zhang, Christian Cosgrove,
  Christopher~D Manning, Christopher Re, Diana Acosta-Navas, Drew~A. Hudson,
  Eric Zelikman, Esin Durmus, Faisal Ladhak, Frieda Rong, Hongyu Ren, Huaxiu
  Yao, Jue WANG, Keshav Santhanam, Laurel Orr, Lucia Zheng, Mert Yuksekgonul,
  Mirac Suzgun, Nathan Kim, Neel Guha, Niladri~S. Chatterji, Omar Khattab,
  Peter Henderson, Qian Huang, Ryan~Andrew Chi, Sang~Michael Xie, Shibani
  Santurkar, Surya Ganguli, Tatsunori Hashimoto, Thomas Icard, Tianyi Zhang,
  Vishrav Chaudhary, William Wang, Xuechen Li, Yifan Mai, Yuhui Zhang, and Yuta
  Koreeda.
\newblock Holistic evaluation of language models.
\newblock \emph{Transactions on Machine Learning Research}, 2023.
\newblock ISSN 2835-8856.
\newblock URL \url{https://openreview.net/forum?id=iO4LZibEqW}.
\newblock Featured Certification, Expert Certification, Outstanding
  Certification.

\bibitem[Lin et~al.(2022)Lin, Hilton, and Evans]{lin2022truthfulqa}
Stephanie Lin, Jacob Hilton, and Owain Evans.
\newblock {T}ruthful{QA}: Measuring how models mimic human falsehoods.
\newblock In Smaranda Muresan, Preslav Nakov, and Aline Villavicencio, editors,
  \emph{Proceedings of the 60th Annual Meeting of the Association for
  Computational Linguistics (Volume 1: Long Papers)}, pages 3214--3252, Dublin,
  Ireland, May 2022. Association for Computational Linguistics.
\newblock \doi{10.18653/v1/2022.acl-long.229}.
\newblock URL \url{https://aclanthology.org/2022.acl-long.229/}.

\bibitem[Lu et~al.(2022)Lu, Bartolo, Moore, Riedel, and
  Stenetorp]{lu2022fantastically}
Yao Lu, Max Bartolo, Alastair Moore, Sebastian Riedel, and Pontus Stenetorp.
\newblock Fantastically ordered prompts and where to find them: Overcoming
  few-shot prompt order sensitivity.
\newblock In Smaranda Muresan, Preslav Nakov, and Aline Villavicencio, editors,
  \emph{Proceedings of the 60th Annual Meeting of the Association for
  Computational Linguistics (Volume 1: Long Papers)}, pages 8086--8098, Dublin,
  Ireland, May 2022. Association for Computational Linguistics.
\newblock \doi{10.18653/v1/2022.acl-long.556}.
\newblock URL \url{https://aclanthology.org/2022.acl-long.556/}.

\bibitem[Mizrahi et~al.(2024)Mizrahi, Kaplan, Malkin, Dror, Shahaf, and
  Stanovsky]{mizrahi2024multiprompt}
Moran Mizrahi, Guy Kaplan, Dan Malkin, Rotem Dror, Dafna Shahaf, and Gabriel
  Stanovsky.
\newblock State of what art? a call for multi-prompt {LLM} evaluation.
\newblock \emph{Transactions of the Association for Computational Linguistics},
  12:\penalty0 933--949, 2024.
\newblock \doi{10.1162/tacl_a_00681}.
\newblock URL \url{https://aclanthology.org/2024.tacl-1.52/}.

\bibitem[Perlitz et~al.(2024)Perlitz, Bandel, Gera, Arviv, Ein-Dor, Shnarch,
  Slonim, Shmueli-Scheuer, and Choshen]{perlitz2024efficient}
Yotam Perlitz, Elron Bandel, Ariel Gera, Ofir Arviv, Liat Ein-Dor, Eyal
  Shnarch, Noam Slonim, Michal Shmueli-Scheuer, and Leshem Choshen.
\newblock Efficient benchmarking (of language models).
\newblock In Kevin Duh, Helena Gomez, and Steven Bethard, editors,
  \emph{Proceedings of the 2024 Conference of the North American Chapter of the
  Association for Computational Linguistics: Human Language Technologies
  (Volume 1: Long Papers)}, pages 2519--2536, Mexico City, Mexico, June 2024.
  Association for Computational Linguistics.
\newblock \doi{10.18653/v1/2024.naacl-long.139}.
\newblock URL \url{https://aclanthology.org/2024.naacl-long.139/}.

\bibitem[Pezeshkpour and Hruschka(2024)]{pezeshkpour2024order}
Pouya Pezeshkpour and Estevam Hruschka.
\newblock Large language models sensitivity to the order of options in
  multiple-choice questions.
\newblock In Kevin Duh, Helena Gomez, and Steven Bethard, editors,
  \emph{Findings of the Association for Computational Linguistics: NAACL 2024},
  pages 2006--2017, Mexico City, Mexico, June 2024. Association for
  Computational Linguistics.
\newblock \doi{10.18653/v1/2024.findings-naacl.130}.
\newblock URL \url{https://aclanthology.org/2024.findings-naacl.130/}.

\bibitem[Plaut et~al.(2025)Plaut, Nguyen, and Trinh]{plaut2024softmax}
Benjamin Plaut, Khanh~Xuan Nguyen, and Tu~Trinh.
\newblock Probabilities of chat {LLM}s are miscalibrated but still predict
  correctness on multiple-choice q\&a.
\newblock \emph{Transactions on Machine Learning Research}, 2025.
\newblock ISSN 2835-8856.
\newblock URL \url{https://openreview.net/forum?id=E6LOh5vz5x}.

\bibitem[Polo et~al.(2024)Polo, Weber, Choshen, Sun, Xu, and
  Yurochkin]{polo2024tinybenchmarks}
Felipe~Maia Polo, Lucas Weber, Leshem Choshen, Yuekai Sun, Gongjun Xu, and
  Mikhail Yurochkin.
\newblock tinybenchmarks: evaluating llms with fewer examples.
\newblock In \emph{Proceedings of the 41st International Conference on Machine
  Learning}, ICML'24. JMLR.org, 2024.

\bibitem[Robinson and Wingate(2023)]{robinson2023mcqa}
Joshua Robinson and David Wingate.
\newblock Leveraging large language models for multiple choice question
  answering.
\newblock In \emph{The Eleventh International Conference on Learning
  Representations}, 2023.
\newblock URL \url{https://openreview.net/forum?id=yKbprarjc5B}.

\bibitem[Rodriguez et~al.(2021)Rodriguez, Barrow, Hoyle, Lalor, Jia, and
  Boyd-Graber]{rodriguez2021evaluation}
Pedro Rodriguez, Joe Barrow, Alexander Hoyle, John~P. Lalor, Robin Jia, and
  Jordan Boyd-Graber.
\newblock Evaluation examples are not equally informative: How should that
  change {NLP} leaderboards?
\newblock In Chengqing Zong, Fei Xia, Wenjie Li, and Roberto Navigli, editors,
  \emph{Proceedings of the 59th Annual Meeting of the Association for
  Computational Linguistics and the 11th International Joint Conference on
  Natural Language Processing (Volume 1: Long Papers)}, pages 4486--4503,
  Online, August 2021. Association for Computational Linguistics.
\newblock \doi{10.18653/v1/2021.acl-long.346}.
\newblock URL \url{https://aclanthology.org/2021.acl-long.346/}.

\bibitem[Sclar et~al.(2024)Sclar, Choi, Tsvetkov, and
  Suhr]{sclar2024formatting}
Melanie Sclar, Yejin Choi, Yulia Tsvetkov, and Alane Suhr.
\newblock Quantifying language models' sensitivity to spurious features in
  prompt design or: How i learned to start worrying about prompt formatting,
  2024.
\newblock URL \url{https://arxiv.org/abs/2310.11324}.

\bibitem[Team et~al.(2025)Team, Kamath, Ferret, Pathak, Vieillard, Merhej,
  Perrin, Matejovicova, Ramé, Rivière, Rouillard, Mesnard, Cideron, bastien
  Grill, Ramos, Yvinec, Casbon, Pot, Penchev, Liu, Visin, Kenealy, Beyer, Zhai,
  Tsitsulin, Busa-Fekete, Feng, Sachdeva, Coleman, Gao, Mustafa, Barr,
  Parisotto, Tian, Eyal, Cherry, Peter, Sinopalnikov, Bhupatiraju, Agarwal,
  Kazemi, Malkin, Kumar, Vilar, Brusilovsky, Luo, Steiner, Friesen, Sharma,
  Sharma, Gilady, Goedeckemeyer, Saade, Feng, Kolesnikov, Bendebury, Abdagic,
  Vadi, György, Pinto, Das, Bapna, Miech, Yang, Paterson, Shenoy, Chakrabarti,
  Piot, Wu, Shahriari, Petrini, Chen, Lan, Choquette-Choo, Carey, Brick,
  Deutsch, Eisenbud, Cattle, Cheng, Paparas, Sreepathihalli, Reid, Tran, Zelle,
  Noland, Huizenga, Kharitonov, Liu, Amirkhanyan, Cameron, Hashemi,
  Klimczak-Plucińska, Singh, Mehta, Lehri, Hazimeh, Ballantyne, Szpektor,
  Nardini, Pouget-Abadie, Chan, Stanton, Wieting, Lai, Orbay, Fernandez,
  Newlan, yeong Ji, Singh, Black, Yu, Hui, Vodrahalli, Greff, Qiu, Valentine,
  Coelho, Ritter, Hoffman, Watson, Chaturvedi, Moynihan, Ma, Babar, Noy, Byrd,
  Roy, Momchev, Chauhan, Sachdeva, Bunyan, Botarda, Caron, Rubenstein,
  Culliton, Schmid, Sessa, Xu, Stanczyk, Tafti, Shivanna, Wu, Pan, Rokni,
  Willoughby, Vallu, Mullins, Jerome, Smoot, Girgin, Iqbal, Reddy, Sheth,
  Põder, Bhatnagar, Panyam, Eiger, Zhang, Liu, Yacovone, Liechty, Kalra, Evci,
  Misra, Roseberry, Feinberg, Kolesnikov, Han, Kwon, Chen, Chow, Zhu, Wei,
  Egyed, Cotruta, Giang, Kirk, Rao, Black, Babar, Lo, Moreira, Martins,
  Sanseviero, Gonzalez, Gleicher, Warkentin, Mirrokni, Senter, Collins, Barral,
  Ghahramani, Hadsell, Matias, Sculley, Petrov, Fiedel, Shazeer, Vinyals, Dean,
  Hassabis, Kavukcuoglu, Farabet, Buchatskaya, Alayrac, Anil, Dmitry, Lepikhin,
  Borgeaud, Bachem, Joulin, Andreev, Hardin, Dadashi, and Hussenot]{gemma3}
Gemma Team, Aishwarya Kamath, Johan Ferret, Shreya Pathak, Nino Vieillard,
  Ramona Merhej, Sarah Perrin, Tatiana Matejovicova, Alexandre Ramé, Morgane
  Rivière, Louis Rouillard, Thomas Mesnard, Geoffrey Cideron, Jean bastien
  Grill, Sabela Ramos, Edouard Yvinec, Michelle Casbon, Etienne Pot, Ivo
  Penchev, Gaël Liu, Francesco Visin, Kathleen Kenealy, Lucas Beyer, Xiaohai
  Zhai, Anton Tsitsulin, Robert Busa-Fekete, Alex Feng, Noveen Sachdeva,
  Benjamin Coleman, Yi~Gao, Basil Mustafa, Iain Barr, Emilio Parisotto, David
  Tian, Matan Eyal, Colin Cherry, Jan-Thorsten Peter, Danila Sinopalnikov,
  Surya Bhupatiraju, Rishabh Agarwal, Mehran Kazemi, Dan Malkin, Ravin Kumar,
  David Vilar, Idan Brusilovsky, Jiaming Luo, Andreas Steiner, Abe Friesen,
  Abhanshu Sharma, Abheesht Sharma, Adi~Mayrav Gilady, Adrian Goedeckemeyer,
  Alaa Saade, Alex Feng, Alexander Kolesnikov, Alexei Bendebury, Alvin Abdagic,
  Amit Vadi, András György, André~Susano Pinto, Anil Das, Ankur Bapna,
  Antoine Miech, Antoine Yang, Antonia Paterson, Ashish Shenoy, Ayan
  Chakrabarti, Bilal Piot, Bo~Wu, Bobak Shahriari, Bryce Petrini, Charlie Chen,
  Charline~Le Lan, Christopher~A. Choquette-Choo, CJ~Carey, Cormac Brick,
  Daniel Deutsch, Danielle Eisenbud, Dee Cattle, Derek Cheng, Dimitris Paparas,
  Divyashree~Shivakumar Sreepathihalli, Doug Reid, Dustin Tran, Dustin Zelle,
  Eric Noland, Erwin Huizenga, Eugene Kharitonov, Frederick Liu, Gagik
  Amirkhanyan, Glenn Cameron, Hadi Hashemi, Hanna Klimczak-Plucińska, Harman
  Singh, Harsh Mehta, Harshal~Tushar Lehri, Hussein Hazimeh, Ian Ballantyne,
  Idan Szpektor, Ivan Nardini, Jean Pouget-Abadie, Jetha Chan, Joe Stanton,
  John Wieting, Jonathan Lai, Jordi Orbay, Joseph Fernandez, Josh Newlan,
  Ju~yeong Ji, Jyotinder Singh, Kat Black, Kathy Yu, Kevin Hui, Kiran
  Vodrahalli, Klaus Greff, Linhai Qiu, Marcella Valentine, Marina Coelho,
  Marvin Ritter, Matt Hoffman, Matthew Watson, Mayank Chaturvedi, Michael
  Moynihan, Min Ma, Nabila Babar, Natasha Noy, Nathan Byrd, Nick Roy, Nikola
  Momchev, Nilay Chauhan, Noveen Sachdeva, Oskar Bunyan, Pankil Botarda, Paul
  Caron, Paul~Kishan Rubenstein, Phil Culliton, Philipp Schmid, Pier~Giuseppe
  Sessa, Pingmei Xu, Piotr Stanczyk, Pouya Tafti, Rakesh Shivanna, Renjie Wu,
  Renke Pan, Reza Rokni, Rob Willoughby, Rohith Vallu, Ryan Mullins, Sammy
  Jerome, Sara Smoot, Sertan Girgin, Shariq Iqbal, Shashir Reddy, Shruti Sheth,
  Siim Põder, Sijal Bhatnagar, Sindhu~Raghuram Panyam, Sivan Eiger, Susan
  Zhang, Tianqi Liu, Trevor Yacovone, Tyler Liechty, Uday Kalra, Utku Evci,
  Vedant Misra, Vincent Roseberry, Vlad Feinberg, Vlad Kolesnikov, Woohyun Han,
  Woosuk Kwon, Xi~Chen, Yinlam Chow, Yuvein Zhu, Zichuan Wei, Zoltan Egyed,
  Victor Cotruta, Minh Giang, Phoebe Kirk, Anand Rao, Kat Black, Nabila Babar,
  Jessica Lo, Erica Moreira, Luiz~Gustavo Martins, Omar Sanseviero, Lucas
  Gonzalez, Zach Gleicher, Tris Warkentin, Vahab Mirrokni, Evan Senter, Eli
  Collins, Joelle Barral, Zoubin Ghahramani, Raia Hadsell, Yossi Matias,
  D.~Sculley, Slav Petrov, Noah Fiedel, Noam Shazeer, Oriol Vinyals, Jeff Dean,
  Demis Hassabis, Koray Kavukcuoglu, Clement Farabet, Elena Buchatskaya,
  Jean-Baptiste Alayrac, Rohan Anil, Dmitry, Lepikhin, Sebastian Borgeaud,
  Olivier Bachem, Armand Joulin, Alek Andreev, Cassidy Hardin, Robert Dadashi,
  and Léonard Hussenot.
\newblock Gemma 3 technical report, 2025.
\newblock URL \url{https://arxiv.org/abs/2503.19786}.

\bibitem[Vania et~al.(2021)Vania, Htut, Huang, Mungra, Pang, Phang, Liu, Cho,
  and Bowman]{vania2021comparing}
Clara Vania, Phu~Mon Htut, William Huang, Dhara Mungra, Richard~Yuanzhe Pang,
  Jason Phang, Haokun Liu, Kyunghyun Cho, and Samuel~R. Bowman.
\newblock Comparing test sets with item response theory.
\newblock In Chengqing Zong, Fei Xia, Wenjie Li, and Roberto Navigli, editors,
  \emph{Proceedings of the 59th Annual Meeting of the Association for
  Computational Linguistics and the 11th International Joint Conference on
  Natural Language Processing (Volume 1: Long Papers)}, pages 1141--1158,
  Online, August 2021. Association for Computational Linguistics.
\newblock \doi{10.18653/v1/2021.acl-long.92}.
\newblock URL \url{https://aclanthology.org/2021.acl-long.92/}.

\bibitem[Yang et~al.(2025)Yang, Li, Yang, Zhang, Hui, Zheng, Yu, Gao, Huang,
  Lv, Zheng, Liu, Zhou, Huang, Hu, Ge, Wei, Lin, Tang, Yang, Tu, Zhang, Yang,
  Yang, Zhou, Zhou, Lin, Dang, Bao, Yang, Yu, Deng, Li, Xue, Li, Zhang, Wang,
  Zhu, Men, Gao, Liu, Luo, Li, Tang, Yin, Ren, Wang, Zhang, Ren, Fan, Su,
  Zhang, Zhang, Wan, Liu, Wang, Cui, Zhang, Zhou, and Qiu]{qwen3}
An~Yang, Anfeng Li, Baosong Yang, Beichen Zhang, Binyuan Hui, Bo~Zheng, Bowen
  Yu, Chang Gao, Chengen Huang, Chenxu Lv, Chujie Zheng, Dayiheng Liu, Fan
  Zhou, Fei Huang, Feng Hu, Hao Ge, Haoran Wei, Huan Lin, Jialong Tang, Jian
  Yang, Jianhong Tu, Jianwei Zhang, Jianxin Yang, Jiaxi Yang, Jing Zhou,
  Jingren Zhou, Junyang Lin, Kai Dang, Keqin Bao, Kexin Yang, Le~Yu, Lianghao
  Deng, Mei Li, Mingfeng Xue, Mingze Li, Pei Zhang, Peng Wang, Qin Zhu, Rui
  Men, Ruize Gao, Shixuan Liu, Shuang Luo, Tianhao Li, Tianyi Tang, Wenbiao
  Yin, Xingzhang Ren, Xinyu Wang, Xinyu Zhang, Xuancheng Ren, Yang Fan, Yang
  Su, Yichang Zhang, Yinger Zhang, Yu~Wan, Yuqiong Liu, Zekun Wang, Zeyu Cui,
  Zhenru Zhang, Zhipeng Zhou, and Zihan Qiu.
\newblock Qwen3 technical report, 2025.
\newblock URL \url{https://arxiv.org/abs/2505.09388}.

\bibitem[Zellers et~al.(2019)Zellers, Holtzman, Bisk, Farhadi, and
  Choi]{zellers2019hellaswag}
Rowan Zellers, Ari Holtzman, Yonatan Bisk, Ali Farhadi, and Yejin Choi.
\newblock {H}ella{S}wag: Can a machine really finish your sentence?
\newblock In Anna Korhonen, David Traum, and Llu{\'i}s M{\`a}rquez, editors,
  \emph{Proceedings of the 57th Annual Meeting of the Association for
  Computational Linguistics}, pages 4791--4800, Florence, Italy, July 2019.
  Association for Computational Linguistics.
\newblock \doi{10.18653/v1/P19-1472}.
\newblock URL \url{https://aclanthology.org/P19-1472/}.

\bibitem[Zhao et~al.(2021)Zhao, Wallace, Feng, Klein, and
  Singh]{zhao2021calibrate}
Tony~Z. Zhao, Eric Wallace, Shi Feng, Dan Klein, and Sameer Singh.
\newblock Calibrate before use: Improving few-shot performance of language
  models, 2021.
\newblock URL \url{https://arxiv.org/abs/2102.09690}.

\end{thebibliography}

\appendix

\section{Reproducibility}
\label{app:repro}

Every number in this paper is a deterministic function of the released per-item records and a single seed, 1234. This appendix gives the item selection, the 26 configurations, the decoding parameters, and every step of the analysis that consumes randomness. It is written so that a reader can rebuild the grid from the benchmarks themselves, not only rerun our script on our records.

\subsection{Released artifacts}
\label{app:artifacts}

The release holds 48 record files, one per benchmark and model, in JSONL form. Each line is one item and carries the benchmark, the item identifier, the question text, the gold index, and a \texttt{bits} map from each of the 26 configuration keys to a correctness bit. Two further per-item fields carry the raw material for the extensions proposed in Section~\ref{sec:future}. The margin between the top two options under the plain cloze scorer is the input a per-item flip predictor would need, and the count of generation configurations under which no answer could be parsed is what separates a wrong answer from an unparseable one. The release also holds the configuration legend, which names all 26 keys and the permutation attached to each, and the analysis script that produces \texttt{analysis.json} and the four figures. Regenerating every number takes about ten seconds on a CPU and needs no GPU.

\subsection{Item selection}
\label{app:items}

Items are drawn once, with seed 1234, and the identical set goes to all 12 models. For MMLU we shuffle the test split and take the first 1{,}005 items. For ARC, HellaSwag, and TruthfulQA we build a four-option pool and shuffle it with the same seed. The first 5 items of each shuffled benchmark are reserved as a few-shot pool and are never evaluated. The grid is zero-shot, so that pool goes unused, but we exclude those items anyway to keep the item set identical to the one a few-shot version of the protocol would use. The remaining items give 1{,}000 each from ARC, HellaSwag, and MMLU. TruthfulQA yields 679, because its four-option pool holds 684 items in total.

Two construction details matter for reading the reference leaderboard. ARC contributes only its questions that carry exactly four choices, so items with three or five choices are dropped rather than padded. TruthfulQA is not natively four-option: its multiple-choice split gives one correct answer and a variable number of distractors, and we assemble each item as the correct answer followed by the first three distractors. The gold answer therefore occupies the first position in that benchmark's own option order. Five of the six orderings move it, but the reference configuration uses the identity ordering and so inherits it, which makes the reference answer key uniformly the first option on those 679 items. Section~\ref{app:loo} recomputes every headline number with TruthfulQA removed.

\subsection{The 26 configurations}
\label{app:configs}

A configuration is a triple of scoring method, prompt template, and option ordering. The four generation templates crossed with the six orderings give 24 configurations, and the two likelihood cloze stems, which are order invariant because each option is scored on its own, bring the total to 26. Table~\ref{tab:perms} lists the orderings and Table~\ref{tab:templates} the templates. The reference configuration is generation with the \texttt{letter\_plain} template under the identity ordering.

\begin{table}[ht]
  \caption{The six option orderings. Entry $j$ gives the index of the original option shown in display slot $j$, so \texttt{p0} is the identity and leaves the benchmark's own order intact. The five non-identity orderings are a fixed draw from the 24 permutations of four options, taken under a separate constant seed (99) so that the same five appear in every run and for every model. A model's prediction is mapped back through the permutation before it is scored, so a correct answer counts as correct whatever slot it was shown in.}
  \label{tab:perms}
  \begin{tabular}{@{}lccccl@{}}
    \toprule
    Ordering & Slot A & Slot B & Slot C & Slot D & \\
    \midrule
    \texttt{p0} & 0 & 1 & 2 & 3 & identity; used by the reference configuration \\
    \texttt{p1} & 1 & 3 & 2 & 0 & \\
    \texttt{p2} & 0 & 1 & 3 & 2 & \\
    \texttt{p3} & 0 & 2 & 3 & 1 & \\
    \texttt{p4} & 3 & 0 & 2 & 1 & \\
    \texttt{p5} & 2 & 3 & 1 & 0 & \\
    \bottomrule
  \end{tabular}
\end{table}

\begin{table}[ht]
  \caption{The six prompt templates. Newlines are written \texttt{\textbackslash n}, and the option block renders the four options one per line in the order fixed by the configuration's permutation, with the label style shown. The four generation templates are crossed with the six orderings. The two cloze stems carry no option labels, since each option is scored as a continuation of the stem, so the ordering axis cannot act on them.}
  \label{tab:templates}
  \small
  \begin{tabular}{@{}llp{0.46\linewidth}@{}}
    \toprule
    Template & Labels & Text \\
    \midrule
    \multicolumn{3}{@{}l}{\textit{Generation: the model emits an answer and a regular expression reads it}} \\
    \texttt{letter\_plain} & \texttt{A.} & \texttt{Question: [q]\textbackslash n[options]\textbackslash nAnswer with the letter only:} \\
    \texttt{letter\_paren} & \texttt{(A)} & \texttt{[q]\textbackslash n[options]\textbackslash nThe correct option is} \\
    \texttt{digit\_labels} & \texttt{1.} & \texttt{Question: [q]\textbackslash n[options]\textbackslash nAnswer (1-4):} \\
    \texttt{instruction} & \texttt{A)} & \texttt{Read carefully and pick the single best choice.\textbackslash n[q]\textbackslash n[options]\textbackslash nYour choice:} \\
    \midrule
    \multicolumn{3}{@{}l}{\textit{Likelihood: each option is scored as a continuation and the argmax wins}} \\
    \texttt{cloze\_plain} & none & \texttt{[q]\textbackslash nAnswer:} \\
    \texttt{cloze\_question} & none & \texttt{Question: [q]\textbackslash nAnswer:} \\
    \bottomrule
  \end{tabular}
\end{table}

\textbf{Parsing.} The letter templates are read with the regular expression \texttt{\textbackslash b([A-D])\textbackslash b} against the uppercased output, and the digit template with \texttt{\textbackslash b([1-4])\textbackslash b}. An output that matches neither is scored wrong and counted as unparsed. Both parsers are exact rather than fuzzy, which is deliberate. A fifth template that asks the model to restate the answer text exists in our codebase and is excluded from the grid, because matching free text back to an option requires a similarity heuristic, and parser noise would then be indistinguishable from harness fragility. Every configuration in the grid resolves its answer by an exact match or scores the item wrong.

\subsection{Inference and decoding}
\label{app:inference}

Inference runs under vLLM \citep{kwon2023vllm} on four 48-gigabyte A40 GPUs, one model per process so that each model's weights are freed before the next loads. Models at 70 billion parameters and the mixture-of-expert models are sharded across two cards, and the rest fit on one. Generation is greedy, with temperature 0, top-$p$ 1.0, a seed of 1234, and a cap of 8 new tokens, which suffices for a letter or a digit. Temperature 0 makes the seed inert for generation, and we set it anyway so that a reader who raises the temperature inherits a seeded run. The likelihood path scores each option as a continuation of the cloze stem and takes the mean per-token log-probability, so a long option is not penalized for its length, and selects the option with the highest value. That path has no sampling step at all. Prompts are zero-shot throughout.

\subsection{Every step that consumes randomness}
\label{app:seeds}

Table~\ref{tab:seeds} lists them. There are seven, and six affect a reported number, since greedy decoding leaves the generation seed inert. Within the compression analysis a single generator is drawn from once and consumed in a fixed order, first for the correlation bootstrap, then for the tie-breaks, then for the random subsets, so the analysis is a deterministic function of its seed rather than of the order in which a reader calls its parts.

\begin{table}[ht]
  \caption{Every step in the pipeline that consumes randomness, and what it decides. Nothing else in the pipeline is stochastic: the correctness tensor itself is produced by greedy decoding and by likelihood scoring, both of which are deterministic given the weights.}
  \label{tab:seeds}
  \small
  \begin{tabular}{@{}llll@{}}
    \toprule
    Step & Seed & Draws & What it decides \\
    \midrule
    Item sampling                & 1234 & once per benchmark & which items are evaluated \\
    Option orderings             & 99   & once               & the five non-identity permutations \\
    Generation                   & 1234 & per call           & nothing, since the temperature is 0 \\
    Bootstrap, robust-item gap   & 1234 & 10{,}000           & the intervals behind Table~\ref{tab:pairs} \\
    Bootstrap, correlation       & 1234 & 2{,}000            & the interval on 0.28, not the 0.28 \\
    Discrimination tie-break     & 1234 & 200                & which 64 items complete the top-100 subset \\
    Random-subset baseline       & 1234 & 200                & the 5.1 champion baseline \\
    \bottomrule
  \end{tabular}
\end{table}

\subsection{The discrimination cutoff is tied}
\label{app:tiebreak}

This is the one place in the analysis where a seed changes an answer, and it is worth stating exactly why. Item discrimination is the point-biserial correlation between an item's reference-configuration correctness across the 12 models and those models' reference accuracies. Estimated over a pool of 12, it is a correlation between two 12-vectors, one of them binary, so it can take only a few hundred distinct values however many items there are. Across our 3{,}679 items it takes 720. At the rank-100 cutoff the discrimination value is 0.872, and 36 items lie strictly above it while 102 sit exactly on it. A top-100 subset therefore contains 36 items chosen by the data and 64 chosen by whatever breaks the tie. An ordinary sort would fill those 64 slots by position in the input array, which is arbitrary, undocumented, and not stable across implementations. An earlier version of this analysis did precisely that and reported a champion count that we could not reproduce.

We do not report one draw. We sample 200 tie-breaks from a seeded generator and report the distribution of the champion count over them. Under seed 1234 the counts are 4 once, 5 in 85 draws, 6 in 94, and 7 in 20, giving a mean of 5.67 and a range of 4 to 7. Table~\ref{tab:tiebreak} varies the seed itself. The mean champion count on the discriminative subset moves between 5.59 and 5.72, the random-subset baseline between 4.80 and 5.08, and the discriminative subset exceeds the random baseline under every seed we tried. The seed we report, 1234, gives the smallest gap of the eight, so the number in Section~\ref{sec:compression} is the least favorable of them to our claim. The correlation between discrimination and fragility carries no seed dependence at all: its point estimate, 0.2767, is computed on the full item set and is identical under every seed, and only the width of its interval comes from the bootstrap.

\begin{table}[ht]
  \caption{Seed sensitivity of the compression statistics. Each row reruns the analysis end to end under a different seed. The champion count on the discriminative subset is a mean over 200 tie-breaks of the 102 items on the cutoff. The random baseline is a mean over 200 subsets of 100 items drawn without replacement. The discriminative subset yields more champions than the random subset under every seed, and both yield more than the full benchmark, which yields 4 under all of them.}
  \label{tab:tiebreak}
  \begin{tabular}{@{}lcccc@{}}
    \toprule
    Seed & corr(disc, fragility) & Discriminative 100 & Random 100 & Full benchmark \\
    \midrule
    1234 (reported) & 0.2767 & 5.67 & 5.08 & 4 \\
    1               & 0.2767 & 5.67 & 4.81 & 4 \\
    2               & 0.2767 & 5.68 & 4.93 & 4 \\
    7               & 0.2767 & 5.68 & 4.85 & 4 \\
    42              & 0.2767 & 5.70 & 4.80 & 4 \\
    99              & 0.2767 & 5.67 & 4.89 & 4 \\
    2024            & 0.2767 & 5.72 & 4.98 & 4 \\
    31337           & 0.2767 & 5.59 & 4.88 & 4 \\
    \bottomrule
  \end{tabular}
\end{table}

\textbf{Why the champion rule counts a single winner.} On a subset of $k$ items, accuracy is quantized to multiples of $1/k$, so two models tie at the top far more often on 100 items than on 3{,}679. We count one champion per configuration, breaking a top tie by a fixed model order. Counting every tied model as a champion instead would raise the champion count on small subsets for reasons of arithmetic granularity rather than harness fragility, and it would inflate the very comparison this section rests on. The single-winner rule is the conservative choice and we keep it.

\subsection{Champion margins}
\label{app:margins}

A champion count treats rank one as a fact, but rank one is produced by a sort, and a sort will hand first place to a model that leads by a single item. Table~\ref{tab:margins} therefore gives the winner, the runner-up, and the margin for every one of the 26 configurations, so that a reader can see which first places are verdicts and which are coin flips.

Two things follow from it. The first is that the major champions are not close calls: gemma4-31b takes its 16 configurations by a median of 181.5 items and a minimum of 7, and qwen3-14b takes its 8 by a median of 98.5 items and a minimum of 10. The second is that the split between them is drawn by the prompt template. Every configuration of the plain-letter and parenthesized-letter templates goes to gemma4-31b, and every digit-labeled one goes to qwen3-14b, by margins of 91 items or more. The instruction-phrased block, which also labels its options with letters, is the only generation block the two contest, and there they trade first place as the option order changes, with margins of 7 to 27 items.

The two likelihood configurations are the only ones in the grid whose winner is decided within a handful of items. llama3.3-70b takes one by 1 item and mixtral-8x7b takes the other by 3. We report both as champions, since both attain rank one under a configuration we would defend, and we say plainly that the ordering between those two is not a finding. What is a finding is that the pair at the top under likelihood scoring, second and eleventh under the reference harness, is not the pair at the top under generation.

\begin{table}[ht]
  \caption{The winner of every configuration and its margin over the runner-up, in accuracy and in items out of 3{,}679. The 24 generation configurations are decided by margins of 7 to 292 items. The 2 likelihood configurations are decided by 1 item and 3 items, and are the only first places in the grid that a handful of items would overturn. The letter blocks and the digit block are unanimous, which is the split the champion claim rests on.}
  \label{tab:margins}
  \small
  \begin{tabular}{@{}llrlrr@{}}
    \toprule
    Configuration & Winner & Acc. & Runner-up & Margin & Items \\
    \midrule
    \texttt{gen|letter\_plain|p0} & gemma4-31b & 0.8796 & llama3.3-70b & 0.0478 & 176 \\
    \texttt{gen|letter\_plain|p1} & gemma4-31b & 0.8820 & qwen3-30b-a3b & 0.0470 & 173 \\
    \texttt{gen|letter\_plain|p2} & gemma4-31b & 0.8804 & gemma4-26b-a4b & 0.0560 & 206 \\
    \texttt{gen|letter\_plain|p3} & gemma4-31b & 0.8731 & llama3.3-70b & 0.0468 & 172 \\
    \texttt{gen|letter\_plain|p4} & gemma4-31b & 0.8834 & gemma4-26b-a4b & 0.0429 & 158 \\
    \texttt{gen|letter\_plain|p5} & gemma4-31b & 0.8856 & qwen3-30b-a3b & 0.0546 & 201 \\
    \texttt{gen|letter\_paren|p0} & gemma4-31b & 0.8230 & qwen3-32b & 0.0652 & 240 \\
    \texttt{gen|letter\_paren|p1} & gemma4-31b & 0.8247 & qwen3-32b & 0.0508 & 187 \\
    \texttt{gen|letter\_paren|p2} & gemma4-31b & 0.8190 & qwen3-32b & 0.0587 & 216 \\
    \texttt{gen|letter\_paren|p3} & gemma4-31b & 0.8146 & qwen3-32b & 0.0546 & 201 \\
    \texttt{gen|letter\_paren|p4} & gemma4-31b & 0.8203 & qwen3-32b & 0.0652 & 240 \\
    \texttt{gen|letter\_paren|p5} & gemma4-31b & 0.8247 & qwen3-32b & 0.0565 & 208 \\
    \texttt{gen|digit\_labels|p0} & qwen3-14b & 0.6268 & qwen3-32b & 0.0247 & 91 \\
    \texttt{gen|digit\_labels|p1} & qwen3-14b & 0.6222 & qwen3-30b-a3b & 0.0794 & 292 \\
    \texttt{gen|digit\_labels|p2} & qwen3-14b & 0.6276 & qwen3-32b & 0.0272 & 100 \\
    \texttt{gen|digit\_labels|p3} & qwen3-14b & 0.6203 & qwen3-32b & 0.0264 & 97 \\
    \texttt{gen|digit\_labels|p4} & qwen3-14b & 0.6013 & llama3.3-70b & 0.0767 & 282 \\
    \texttt{gen|digit\_labels|p5} & qwen3-14b & 0.6099 & gemma4-12b & 0.0769 & 283 \\
    \texttt{gen|instruction|p0} & gemma4-31b & 0.7513 & qwen3-14b & 0.0038 & 14 \\
    \texttt{gen|instruction|p1} & qwen3-14b & 0.7521 & gemma4-31b & 0.0027 & 10 \\
    \texttt{gen|instruction|p2} & gemma4-31b & 0.7518 & qwen3-14b & 0.0068 & 25 \\
    \texttt{gen|instruction|p3} & gemma4-31b & 0.7442 & qwen3-14b & 0.0073 & 27 \\
    \texttt{gen|instruction|p4} & gemma4-31b & 0.7418 & qwen3-14b & 0.0019 & 7 \\
    \texttt{gen|instruction|p5} & qwen3-14b & 0.7543 & gemma4-31b & 0.0060 & 22 \\
    \midrule
    \texttt{ll|cloze\_plain} & llama3.3-70b & 0.6099 & mixtral-8x7b & 0.0003 & 1 \\
    \texttt{ll|cloze\_question} & mixtral-8x7b & 0.6053 & llama3.3-70b & 0.0008 & 3 \\
    \bottomrule
  \end{tabular}
\end{table}

\subsection{Determinism check}
\label{app:determinism}

The analysis script run twice on the same records produces byte-identical output. The pure functions underneath it, which are the item classification, the accuracy band, the axis-robust accuracy, the pair decomposition, and the bootstrap, carry 54 unit tests.

\subsection{Leaving TruthfulQA out}
\label{app:loo}

Section~\ref{app:items} noted that TruthfulQA's four-option items are assembled with the gold answer first, which the reference configuration inherits. A reader is entitled to ask whether that artifact produces our results. It does not, and the direction of the effect runs against us: on TruthfulQA, 10 of the 12 models score \textit{lower} under the reference configuration, where the answer is always the first option, than under the first permuted ordering, where it is the last. Only llama3.2-3b gains materially, by 0.190, and it places last on the leaderboard regardless. The reference configuration is thus harder than a permuted one for most of the roster on those items rather than easier.

Table~\ref{tab:loo} recomputes every headline number on the other three benchmarks. All three claims hold, and two strengthen. Every adjacent pair becomes order-manufactured rather than 10 of 11, and the correlation between discrimination and fragility rises from 0.28 to 0.32. The fragile share of the gap exceeds 1 on the reduced set, which is not a rounding artifact. For the qwen3-14b and gemma4-12b pair the jointly robust items favor the \textit{lower}-ranked model by one item, so the robust part of the gap is negative and the fragile part more than accounts for the whole of it. Eight of the 11 pairs have a robust component of exactly zero.

Two secondary numbers do move, and we report them rather than bury them. The champion count falls from 4 to 3, because mixtral-8x7b scores well above its own leaderboard position on TruthfulQA and loses the single configuration under which it takes rank one. The minimum Kendall correlation rises from $-0.09$ to $+0.03$, so the configuration that produces an ordering unrelated to the reference is one whose behavior depends on TruthfulQA being in the pool. The existence claim in Section~\ref{sec:champions} is therefore a claim about the item set we evaluate. Three defensible champions rather than four would not change its force, but we do not claim the count is invariant to the benchmark mix.

\begin{table}[ht]
  \caption{Every headline number recomputed with TruthfulQA removed. The three claims of the paper hold on either item set. The two quantities that move are the champion count and the minimum Kendall correlation, both of which fall on the reduced set. The mean fragile share exceeds 1 without TruthfulQA because one pair's jointly robust items favor the lower-ranked model.}
  \label{tab:loo}
  \small
  \begin{tabular}{@{}lcc@{}}
    \toprule
    Quantity & 4 benchmarks (reported) & 3 benchmarks (no TruthfulQA) \\
    \midrule
    Items per model                                  & 3{,}679 & 3{,}000 \\
    Mean config-lucky fraction                       & 0.853 & 0.859 \\
    Adjacent pairs whose order is manufactured       & 10 of 11 & 11 of 11 \\
    Mean fragile share of gap                        & 0.957 & 1.004 \\
    Distinct champions                               & 4 & 3 \\
    Models reaching the top three                    & 8 & 9 \\
    corr(discrimination, fragility)                  & 0.28 [0.25, 0.30] & 0.32 [0.29, 0.35] \\
    Mean fragility, 100 most discriminative items    & 0.960 & 0.968 \\
    Mean fragility, all items                        & 0.851 & 0.863 \\
    Axis-robust accuracy, option order               & 0.598 & 0.605 \\
    Axis-robust accuracy, prompt format              & 0.396 & 0.367 \\
    Axis-robust accuracy, scoring method             & 0.314 & 0.328 \\
    Kendall $\tau$ against the reference, mean       & 0.42 & 0.41 \\
    Kendall $\tau$ against the reference, minimum    & $-0.09$ & $+0.03$ \\
    \bottomrule
  \end{tabular}
\end{table}

\end{document}